\documentclass[prd,aps,amsmath,twocolumn,superscriptaddress,amssymb,reprint,nofootinbib,preprintnumbers,longbibliography]{revtex4-2}
\usepackage{amsmath, amsthm}
\usepackage{amssymb}
\usepackage{graphicx}
\usepackage{xcolor}
\usepackage{float}
\graphicspath{{images/}}
\usepackage{hyperref}
\hypersetup{
pdfborder={0 0 1},
	colorlinks=true,
	linkcolor=blue,
	citecolor=blue,
	urlcolor=blue,
}

\usepackage{amsmath,amsfonts,bm,xspace}

\def\1{\bm{1}}

\DeclareMathAlphabet{\mathsfit}{\encodingdefault}{\sfdefault}{m}{sl}
\SetMathAlphabet{\mathsfit}{bold}{\encodingdefault}{\sfdefault}{bx}{n}

\def\beq{\begin{equation}\displaystyle}
\def\eeq{\end{equation}}
\def\bea{\begin{eqnarray}\displaystyle} 
\def\eea{\end{eqnarray}}

\def\({\left(}
\def\){\right)}
\def\[{\left[}
\def\]{\right]}

\newcommand{\methodname}[0]{{SIGMA}\xspace}

\definecolor{lightblue}{rgb}{0.1, 0.5, 1.0}
\usepackage{tikz}
\usetikzlibrary{shapes.geometric, arrows.meta, calc}
\usepackage{xcolor}

\definecolor{boxgray}{RGB}{220,220,220}
\definecolor{bluelight}{RGB}{230,241,251}
\definecolor{blueborder}{RGB}{17,85,204}
\definecolor{bluetext}{RGB}{17,85,204}
\definecolor{pinklight}{RGB}{249,226,226}
\definecolor{pinkborder}{RGB}{204,0,0}
\definecolor{redtext}{RGB}{204,0,0}

\begin{document}
\renewcommand{\bibsection}{\section*{References}}
\renewcommand{\footnoterule}{\vspace*{4pt}}

\title{An end-to-end framework for uncertainty-aware
validation of generative models}

\author{Giada Badaracco}
\email{gbadaracco@phys.ethz.ch}
\affiliation{ETH Zürich, Zürich, CH}
\affiliation{NSF AI Institute for Artificial Intelligence and Fundamental Interactions, Cambridge, MA, USA}

\author{Sean Benevedes}
\affiliation{School of Physics, Georgia Institute of Technology, Atlanta, GA, USA}
\affiliation{MIT Center for Theoretical Physics -- a Leinweber Institute, Cambridge, MA, USA}
\affiliation{NSF AI Institute for Artificial Intelligence and Fundamental Interactions, Cambridge, MA, USA}

\author{Christina Reissel}
\affiliation{NSF AI Institute for Artificial Intelligence and Fundamental Interactions, Cambridge, MA, USA}
\affiliation{MIT Laboratory for Nuclear Science, Cambridge, MA, USA}

\author{Philip Harris}
\affiliation{NSF AI Institute for Artificial Intelligence and Fundamental Interactions, Cambridge, MA, USA}
\affiliation{MIT Laboratory for Nuclear Science, Cambridge, MA, USA}

\author{Gaia Grosso}
\affiliation{NSF AI Institute for Artificial Intelligence and Fundamental Interactions, Cambridge, MA, USA}
\affiliation{MIT Laboratory for Nuclear Science, Cambridge, MA, USA}

\markboth{}{An end-to-end framework for uncertainty-aware validation of generative models}


\begin{abstract}
Density estimation with generative models is a common task in the physical sciences, for example in background estimation in particle physics. Existing machine learning techniques, however, lack methods to quantify uncertainties that can be propagated into a statistical test. That limits the validation of generative models to standard goodness-of-fit tests, in which the generative model is treated as exact (point-null hypothesis).
But a model fitted to a finite training set is never exact, so if its uncertainties are neglected, any sufficiently powerful goodness-of-fit test will eventually reject it once the test set is large enough to resolve this estimation error. 
We propose \methodname, an end-to-end framework combining density estimation with principled uncertainty quantification.
We use Gaussian mixture models, for which uncertainties are tractable, and further construct a new goodness-of-fit test in which these uncertainties are profiled as nuisance parameters to account for the statistical error of the fit, promoting the conventional point-null test to a composite one.
A rejection then signals genuine mismodeling rather than a test set large enough to resolve the error of the fit, provided the propagated uncertainties faithfully cover that error.
We demonstrate the framework on a controlled two-dimensional toy model and on a collider physics benchmark modeling the QCD background process.
With uncertainties included, the same model can be trusted on test sets up to an order of magnitude larger than the standard test allows. 
Our work thus provides a statistically grounded basis for a robust use of generative models in the physical sciences.

\end{abstract}
\maketitle
\tableofcontents

\section{Introduction}\label{sec:intro}
\begin{figure}[htbp]
\centering
\includegraphics[scale=0.9]{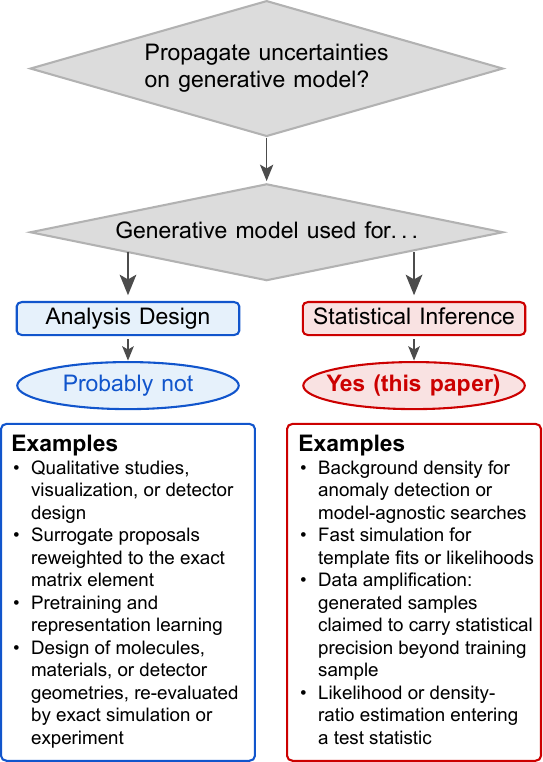}
\caption{\textbf{When uncertainty quantification matters.} Representative uses of generative models in the physical sciences, classified by whether the error of the fit should be propagated into a quantitative claim. For the uses on the left (blue), a wrong model costs efficiency or statistical power but not validity of the claim, and qualitative validation suffices. For the uses on the right (red), the fitted density is the reference model of a statistical inference: its error is indistinguishable from a signal, so it must be propagated into the test (Sec.~\ref{sec:method}).}
\label{fig:when}
\end{figure}

Many scientific analyses compare observed data with a reference model in search of significant deviations, which may signal new phenomena. The reliability of the reference model is therefore critical: any mismodeling can be mistaken for a genuine signal and lead to a false discovery. Such reference models are often built from high-fidelity simulations that require substantial computing resources. In particle physics, for example, the background from known processes is modeled by simulating the full chain of hard process, parton showering, and detector response. The demands on the reference model are particularly high in model-agnostic searches, for example anomaly detection~\cite{AnomalyDetection}, where any deviation from it across the full feature space may be reported as a signal.

Generative models have emerged as a way to significantly reduce the cost of such expensive simulations, acting as fast surrogates (or even replacements)~\cite{MLreview,DiSipio:2019imz,Hallin:2021wme,Raine:2022hht,Das:2024eie,Krause:2023uww} or as learned likelihoods in simulation-based inference~\cite{SBI,MiningGold}.
Like any fitted model, however, generative models carry an inherent error due to the finite size of the training set. Generated events are moreover not independent draws from the true density: they all come from a single model (or ensemble of models) fitted to a finite sample, whose fluctuations are imprinted in the model. No amount of generated data can therefore add information beyond the training sample~\cite{Prasanth}.
In practice, any trained model will eventually fail a statistical test if it is checked against enough data. This happens even when the model is, in principle, capable of describing the data perfectly (Sec.~\ref{sec:motivation}). 
Whether this failure matters depends on the use case in the analysis workflow~\cite{Grosso:2026fgo}. A few representative examples are listed in Fig.~\ref{fig:when}. Errors can be tolerable when the result does not impact the downstream task; this includes examples when the model's output is further validated by exact downstream checks, or when only qualitative fidelity is required, as is the case in exploratory studies or pretraining. But when the fitted density serves as the reference model of a statistical test, as a background estimate, a null hypothesis, or a calibration input, its error is indistinguishable from a signal. 
To be used reliably in this role, a generative model must therefore come with uncertainties, and these must enter the statistical test itself.

For models with few parameters, estimating uncertainties follows standard procedures. However, for modern density estimators, particularly machine learning driven generative models, with high-dimensional inputs and large numbers of non-identifiable parameters, the problem becomes technically challenging, and attributing robust uncertainties to them remains an active area of research.

To address this challenge, we propose \methodname (Score-Induced Gaussian Mixture Approximation), a generative model whose uncertainties are tractable and can be propagated into a statistical test. To validate it, we further introduce an uncertainty-aware goodness-of-fit test that incorporates the derived uncertainties.

\methodname models the density with a Gaussian mixture model and describes its uncertainties in the tangent space spanned by its score functions around the best fit of the density. Unlike for neural networks, this simple Ansatz makes the asymptotic maximum-likelihood covariance of the parameters tractable. The test then propagates these uncertainties by promoting the conventional point-null Neyman--Pearson likelihood-ratio test~\cite{DAgnolo:2018cun,LearningMultivariateNP,NeymanPearsonGoF} (also referred to as simple hypothesis) to a composite one~\cite{dAgnolo:2021aun}, in which the nuisance parameters encoding the uncertainties are profiled, as is common practice in LHC searches~\cite{Cowan:2010js}. Unlike in searches, the nuisance parameters are not systematic uncertainties of the experiment, but the statistical uncertainties of the fitted generative model itself. Because they enter the test in the same standard form, conventional systematic uncertainties can be profiled alongside them, so that a single composite test accounts for both the error of the fit and the systematics of the analysis. 
The test outcome is therefore determined by the compatibility of the data with the model within its uncertainties, rather than driven by the fit error.

Calibrated with pseudo-experiments drawn from the composite-null, the test retains its nominal false-positive rate even when the number of test events exceeds the training sample size, provided that the uncertainties faithfully cover the error of the fit.

The paper is organized as follows. In Sec.~\ref{sec:motivation}, we motivate the problem with a controlled one-dimensional example. Section~\ref{sec:related} reviews related work. In Sec.~\ref{sec:method}, we describe the components of our algorithm, from the density model to the uncertainty quantification and the design of the goodness-of-fit test. Section~\ref{sec:exp} applies the method to a controlled two-dimensional toy study, where the true density is known, and to the QCD background in a learned embedding of particle jets, a setting relevant for anomaly detection searches. We conclude in Sec.~\ref{sec:conclusion} and outline future research directions. 

In App.~\ref{app:KL_discussion} we further discuss the intrinsic limits of generative models and standard goodness-of-fit tests.
Appendix~\ref{app:nf} explores a related approach to quantifying and propagating model uncertainties with ensembles of neural-network density estimators, and finds that the resulting uncertainties are insufficient for the composite-null goodness-of-fit test to close on large observed datasets. Appendix~\ref{app:implementation} contains technical details of the experiments and studies the effects of varying a hyperparameter, App.~\ref{app:null-distribution} derives the asymptotic distribution of the composite-null test statistic under the null hypothesis, and App.~\ref{app:resolution} studies the choice of $\sigma$, another hyperparameter in the goodness-of-fit test.

\section{Powerful tests reject every generative model}\label{sec:motivation}
\begin{figure*}[t]
    \includegraphics[width=\linewidth]{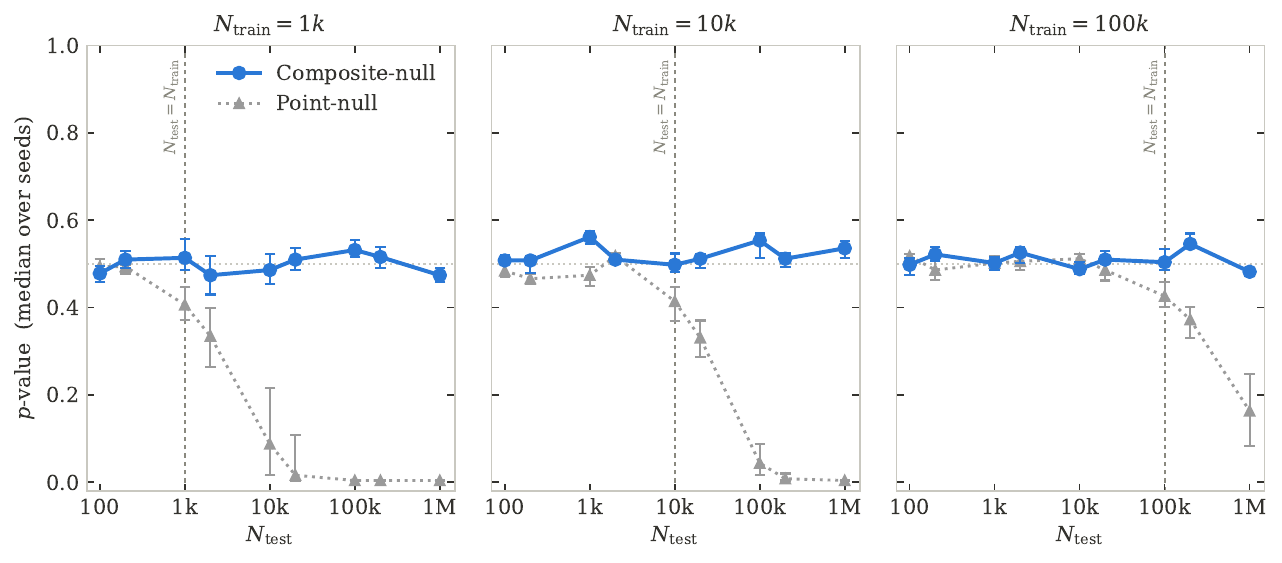}
  \caption{\textbf{Impact of fit uncertainties on goodness-of-fit tests of generative models.}
A Gaussian $\mathcal{N}(\mu,\sigma)$ is fitted by maximum likelihood to $N_{\rm train}$ samples drawn from $\mathcal{N}(0,1)$, with $N_{\rm train}=10^3$, $10^4$ and $10^5$ (left to right). The fitted model is then tested on $N_{\rm test}$ fresh samples from the ground truth, using either the point-null (gray), which takes the fit as exact, or the Gaussian composite-null (blue), which includes the fit uncertainties. For each $N_{\rm train}$, the fit is repeated with 10 independent training samples and each fit is tested on 100 test samples. Markers show the median $p$-value across trainings, and error bars show the 68\% spread of the per-training medians. If the tested null hypothesis holds, $p$-values are uniformly distributed with median $0.5$ (horizontal dotted line).  
Although the model family contains the ground truth, the fitted model never coincides with it exactly. The point-null, which ignores this estimation error, is therefore eventually rejected once $N_{\rm test}\gtrsim N_{\rm train}$ (vertical dashed line). The composite-null accounts for it and remains satisfied up to the largest test size probed, $N_{\rm test}=10^6$, i.e. between $10$ and $10^3$ times the training-set size.}
  \label{fig:pvalue-degradation}
\end{figure*}

A generative model with fitted parameters $\hat{\theta}$ only ever provides an \emph{approximation} $p_{\hat\theta}(x)$ of the true data distribution $p_{0}(x)$. 
Following Refs.~\citep{Haussmann:2026,Jimenez:2026}, the discrepancy has two classes of sources, both epistemic, i.e.\ uncertainties about the model rather than about the intrinsic randomness of the data:
\begin{itemize}
    \item[i)] The \emph{model bias}, a systematic uncertainty: the model may be unable to represent the true distribution exactly, however well it is trained. This bias depends on the choice of architecture and does not shrink with the size of the training sample. It is distinct from the systematic uncertainties usually considered in physics analyses, such as imperfect knowledge of the detector or of the theory, which concern how well the training data, typically a simulation, describe reality.
    \item[ii)] The \emph{statistical uncertainty of the fit}: the finite training sample limits how precisely the parameters can be determined (data variability), and training may converge to different solutions depending on its initialization (optimization variability). Both vanish in the limit of infinite data and ideal optimization.
\end{itemize}
Our work addresses the statistical uncertainty ii). Even in the best-case scenario in which the model family contains the ground truth, the finite size of the training dataset places a fundamental limit on the accuracy of the fitted density.
This limit has a direct and often underappreciated consequence for the \emph{validation} of generative models via goodness-of-fit tests. 

To understand the effect, we consider the following one-dimensional experiment: 
samples are drawn from a standard normal distribution, $p_{0}=\mathcal{N}(0,1)$, and the generative model is chosen to be perfectly well specified, $p_{\theta}=\mathcal{N}(\mu,\sigma)$ with trainable parameters $\theta=(\mu,\sigma)$.

The parameters are estimated by maximum likelihood optimization on $N_{\text{train}}$ training points, and the fitted model is then compared with independent test samples of increasing size $N_{\text{test}}$ by means of a goodness-of-fit test. 
The fitted model, indistinguishable from the truth at small test sample sizes, is eventually rejected with high significance once $N_{\text{test}}$ grows past $N_{\text{train}}$.
This degradation is not specific to the test choice: any consistent goodness-of-fit test eventually detects the finite-statistics error between $p_{\hat\theta}(x)$ and the true density $p_{0}(x)$, so the $p$-values decrease with $N_\text{test}$ regardless of the test used.
The gray markers in Figure~\ref{fig:pvalue-degradation} show this trend for the Gaussian example and the point-null likelihood-ratio test that will be defined in Sec.~\ref{subsec:gof} (see Eq.~\eqref{eq:tpoint}) calibrated with pseudo-experiments drawn from the fitted model.
This behavior is not a pathology of the example but a generic and quantifiable consequence of information limits~\cite{Prasanth} (see App.~\ref{app:KL_discussion} for a derivation).
How fast the $p$-value curve degrades depends both on the sensitivity of the goodness-of-fit test employed and on the quality of the model design, i.e. how well the model design matches the properties of the true distribution.
Crucially, however, the degradation itself cannot be avoided: no amount of architectural or training improvement makes a density estimated from $N_{\text{train}}$ samples \emph{exact}.
How, then, can generative models be used reliably for science? 
It is worth being precise about what this ``intrinsic'' failure means. 
Standard goodness-of-fit methods test a \emph{point-null hypothesis}: the fitted model is treated as an exact description of the data. 
Since the fitted model differs from the truth with probability one, rejection at large $N_{\text{test}}$ is guaranteed by construction. 
However, this is not the question of scientific interest. 
The relevant question is whether the data are compatible with the generative model \emph{within its statistical uncertainty}, the spread of fits that would result from different training samples and different runs of the optimization.
This is a \emph{composite-null hypothesis}. 
A test that ignores this uncertainty conflates genuine mismodeling with the unavoidable statistical fluctuations of the model fit, and its outcome is driven by the relation between $N_{\text{test}}$ and $N_{\text{train}}$ rather than by the quality of the model.
On the other hand, the bias introduced by an incorrect model is one that uncertainties cannot solve, and needs to be addressed at the level of model design, e.g.\ by replacing parametric background functions that fail as the dataset grows with more flexible Gaussian processes~\citep{Frate:2017mai}.
The statistical uncertainty of the fit, however, can be propagated into the goodness-of-fit test itself, promoting the point-null to a composite one. With the uncertainties included, the median $p$-value remains close to its nominal value of 0.5 over the full range of test sizes we probe (blue markers in Fig.~\ref{fig:pvalue-degradation}).

\section{Related work}\label{sec:related}
\paragraph{Generative models and uncertainty quantification.}
Uncertainties on generative models have been estimated with Bayesian approaches~\citep{BayesianFlows, Butter:2026dvv} and Monte Carlo dropout~\citep{gal2016dropout,Dropout}.
While both return a spread of predictions, they do not come in a form that can be written down as a parametrized density function. This makes them incompatible with composite likelihood-ratio tests, which require a parametrized family of null densities.

The frequentist ensembling in Ref.~\citep{Sean} provides one such parametrization, describing the uncertainty through the weights of a fixed ensemble of trained models. Ensembling can be performed for any architecture, but the resultant uncertainties span only the coefficients of the trained models. As a result, the uncertainties cover the true density only under the strong assumption that it lies within the span of the trained ensemble members. Ensembling is also used in experimental searches, where the spread of independently trained models provides the uncertainty of a generative background model in the final fit~\citep{ATLAS:2026ugi}.
Our \methodname approach follows a similar logic but spans the full parameter space of the generative model, weakening the well-specification assumption, at the price of requiring a tractable model likelihood.

\paragraph{Data amplification and the statistical value of generated samples.}
\label{subsec:related-amplification}
A closely related line of work asks whether samples drawn from a generative model carry statistical precision beyond that of the training sample, quantified through an amplification factor~\citep{GANplifying,Calomplification,Bayesiamplification,ForecastingAmplification}. Refs.~\citep{ForecastingAmplification,Bayesiamplification} quantify the uncertainty of the learned densities, but do not propagate any uncertainties to the point-null hypothesis.
Ref.~\citep{Prasanth} establishes that the generated data cannot contain more information than the data on which the model was trained, and Ref.~\citep{InfoLimitAmplification} derives an information-theoretic bound on the achievable gain, posing a theoretical limit to amplification.

We address a prerequisite for trustworthy amplification: whether the fitted density, together with its estimated uncertainty, is compatible with independent data from the true distribution.

\paragraph{Validating generative models.}
Previous work typically validates the learned density through a low-dimensional summary test statistic: quantile occupancies~\citep{GANplifying}, projected Jensen--Shannon
divergences of high-level observables~\citep{Calomplification}, the per-bin coverage of Bayesian uncertainties~\citep{Bayesiamplification}, the Kolmogorov asymptotics of a one-dimensional statistic~\citep{ForecastingAmplification}, or classifier-based scores~\citep{Krause:classifier-metric,Kansal:2022spb}, including the full distribution of classifier weights~\citep{Das:2023ktd}. 
Since all of these tests treat the fitted density as exact, they eventually reject even a model that is as good as its training statistics allow, once $N_{\rm test}$ is large enough to resolve the error of the fit (Sec.~\ref{sec:motivation}).

Two-sample tests for validating generative models have been compared in Refs.~\citep{Grossi:2024axb,Grossi:2025pmm}. The latter finds the Neyman--Pearson test to be overall the most robust, outperforming marginal-based comparisons, at a higher computational cost.
The Neyman--Pearson test with a learned alternative
hypothesis~\citep{DAgnolo:2018cun,LearningMultivariateNP,NeymanPearsonGoF,Bressler:2024wzc} compares the hypotheses in the full feature space, assumes little about the position or shape of a deviation, and returns a $p$-value that is calibrated under the assumption that the reference model is exact. 
While Refs.~\citep{Cappelli:2025myc,SPARKER} applied the Neyman--Pearson test to the validation of generative models in high-dimensional settings, these studies did not account for their uncertainties.
Ref.~\citep{dAgnolo:2021aun} extends the Neyman--Pearson test to include nuisance parameters in the standard LHC form~\citep{Cowan:2010js}. We adopt this structure, but in our case the nuisance parameters are not hand-chosen systematic uncertainties, but the estimation uncertainty of the generative model.

\paragraph{Gaussian mixture models.}
Gaussian mixture models are a classical tool of statistics~\cite{mclachlan2019finite}. Their maximum-likelihood fitting via expectation--maximization (EM)~\cite{dempster1977maximum,mclachlan2008algorithm}, the computation of the observed information matrix~\cite{louis1982finding}, and the asymptotic theory of mixture likelihoods~\cite{redner1984mixture,McLachlanPeel} are all well established. So are the ways mixtures depart from textbook regularity: components can be degenerate or weakly identified, and the model is identifiable only up to relabeling~\cite{yakowitz1968identifiability}. These issues motivate the regularization choices in Sec.~\ref{sec:method}. We claim no novelty on this machinery. Our contribution is to use the tractable score and Fisher information of a fitted mixture to extend the point-null goodness-of-fit test to a composite one, so that the uncertainty in the fit enters the test itself.

\paragraph{Goodness-of-fit with estimated parameters.}
Standard goodness-of-fit tests such as Kolmogorov–Smirnov \cite{massey1951kolmogorov} assume a fully specified null.
Testing a null hypothesis that was itself fitted to data is a classical problem in statistics~\cite{lilliefors1967kolmogorov,durbin1973weak}, and the standard remedy is the parametric bootstrap
~\cite{stute1993bootstrap,babu2004goodness, genest2008validity}. Particle physics knows the same process as calibration with pseudo-experiments, or toy Monte Carlo~\cite{hayes1989application,cowan1998statistical}. Our bootstrap prescription of Sec.~\ref{subsec:tangent-calibration} mirrors this classical procedure, combined with a composite-null test in which the fit uncertainty enters as an explicit nuisance parametrization.

\section{Method}\label{sec:method}
\begin{figure*}[t]
  \centering
  \includegraphics{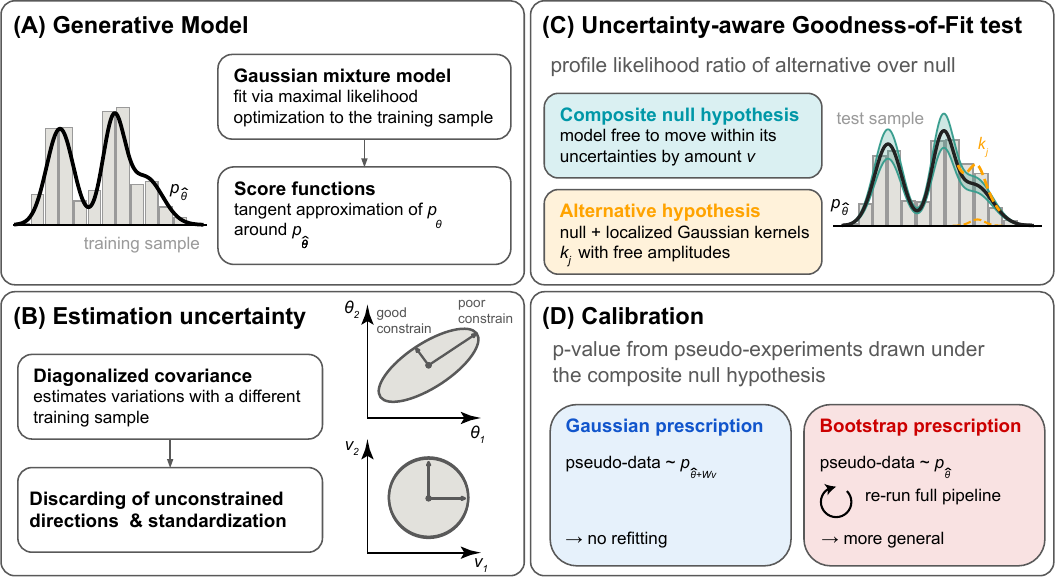}
  \caption{\textbf{Schematic overview of the method.}
\textbf{(A)}~The generative model $p_{\hat\theta}$ is fitted by maximum likelihood and linearized around the best fit, so that parameter shifts act through the fixed score $\psi(x)$.
\textbf{(B)}~The Fisher information sets the uncertainty of the fit, and standardized coordinates measure parameter shifts in units of that uncertainty.
\textbf{(C)}~The test compares a null family $g_v$ (green band), in which the model moves within its uncertainty, with an alternative $h_{v,\alpha}$ (orange dashed) that also switches on localized Gaussian kernels. \textbf{(D)}~The null distribution of the test statistic is obtained from pseudo-experiments following the Gaussian or bootstrap prescription.}
  \label{fig:method-schematic}
\end{figure*}
The method, summarized in Fig.~\ref{fig:method-schematic}, has three parts: a generative model with an unconstrained parametrization (Sec.~\ref{subsec:model}), a standardized description of its estimation uncertainty (Sec.~\ref{subsec:uncertainty}), and a goodness-of-fit test in which the model may move within that uncertainty (Sec.~\ref{subsec:gof}). The calibration of the goodness-of-fit test is discussed in Sec.~\ref{subsec:tangent-calibration}.

\subsection{Generative model}
\label{subsec:model}

The model family should be chosen to satisfy two conditions. First, we find that the uncertainty must be propagated on all of the model's parameters, since all of them fluctuate with the training sample. Appendix~\ref{app:nf} shows that approaches propagating it on only a subset, such as the mixture coefficients of an ensemble of trained normalizing flows, fail. Second, every point of parameter space must define a valid density, so that Gaussian fluctuations of the parameters never leave the model family. Gaussian mixture models (GMMs)~\cite{McLachlanPeel} with a suitable parametrization satisfy both. For $K$ components in $D$ dimensions,
\begin{equation}
  p_\theta(x) = \sum_{k=1}^{K} w_k\, \mathcal{N}(x \mid \mu_k, \Sigma_k),
  \label{eq:gmm}
\end{equation}
where the weights $w_{k}$ are given by softmax logits and the covariances, $\Sigma_{k}$, by Cholesky factors with a log-parametrized diagonal,
\begin{equation}
  w_k = \frac{e^{\ell_k}}{\sum_{j=1}^{K} e^{\ell_j}}, \quad
  \Sigma_k = L_k L_k^\top, \quad
  [L_k]_{ii} = e^{s_{k,i}} .
  \label{eq:gmm-param}
\end{equation}
The logits $\ell_k$, means $\mu_k$, log-diagonals $s_{k,i}$, and off-diagonal entries $[L_k]_{ij}$ ($i>j$) are all unconstrained real numbers, and fixing one of the $\ell_k$ using the $\sum_k w_k=1$ constraint removes the invariance of the softmax under a common shift of the logits. The parameter vector $\theta$ therefore ranges over all of $\mathbb{R}^P$, with
\begin{equation}
  P = K\left(1 + D + \tfrac{D(D+1)}{2}\right) - 1 .
  \label{eq:gmm-dim}
\end{equation}

The parameter count grows as $P \sim K D^2/2$, and the number of components needed to resolve a complex density also tends to grow with dimension. In high dimensions, Gaussian mixtures thus become heavily over-parametrized~\citep{Bouveyron:2014}, and neural density estimators such as normalizing flows have largely replaced them~\citep{NormalizingFlows,MLreview}. In low dimensions, however, they remain expressive~\citep{Menary:2021} and offer what neural estimators lack: a moderate number of parameters, identifiable up to a relabeling of the components~\cite{yakowitz1968identifiability}, with analytic scores from which the uncertainty of the fit can be computed. This makes them well suited to the low-dimensional learned embeddings considered in Sec.~\ref{sec:exp}. An alternative, strongly related approach with an ensemble of independently trained flows, inspired by Ref.~\citep{Sean}, is presented in App.~\ref{app:nf}.

The GMM is fitted by maximum likelihood to a training sample $\{x_i\}_{i=1}^{N_{\rm train}}$ drawn from the unknown true density $p_0$,
\begin{equation}
  \hat\theta = \arg\max_{\theta} \sum_{i=1}^{N_{\rm train}} \log p_\theta(x_i),
  \label{eq:mle}
\end{equation}
using the expectation--maximization (EM) algorithm~\cite{dempster1977maximum}, the standard tool for maximum-likelihood estimation of mixtures~\cite{McLachlanPeel}.
Starting from a $k$-means initialization, EM alternates between computing the posterior probability that each training point belongs to each component (E step) and re-estimating the weights, means, and covariances in closed form with the points weighted by those probabilities (M step); each iteration increases the likelihood, and the algorithm converges to a stationary point. 
The GMM likelihood is highly multimodal, so the resulting $\hat{\theta}$ is in general a local maximum~\cite{redner1984mixture}.

For small displacements $\delta\theta$ around the fit,
\begin{equation}
  \log p_{\hat\theta + \delta\theta}(x) = \log p_{\hat\theta}(x) + \delta\theta^\top \psi(x) + O(\|\delta\theta\|^2),
  \label{eq:tangent-expansion}
\end{equation}
where
\begin{equation}
  \psi_a(x) \equiv \left.\frac{\partial \log p_\theta(x)}{\partial \theta_a}\right|_{\theta = \hat\theta}, \qquad a = 1,\dots,P,
  \label{eq:score}
\end{equation}
is the score function. Once the fit is done, $\psi$ is a fixed function of $x$ (see App.~\ref{app:implementation} for its computation). The test of Sec.~\ref{subsec:gof} is built in this tangent space.

\subsection{Estimation uncertainty}
\label{subsec:uncertainty}

We test the fitted model on an independent test sample $\mathcal{D} = \{x_i\}_{i=1}^{N_{\rm test}} \sim p_0$. The standard approach treats the fit as a point-null hypothesis, $H_0^{\rm point}\!: \mathcal{D} \sim p_{\hat\theta}$. But $\hat\theta$ is itself a random variable, which scatters around the best parameter value by $O(N_{\rm train}^{-1/2})$.  
The modified hypothesis, a composite-null hypothesis, $H_0^{\rm comp}$, asks whether the test data are consistent with $p_\theta$ for some $\theta$ close to $\hat\theta$ within the range of values $\theta$ could take due to random variation in the training sample. 
The test statistic of Sec.~\ref{subsec:gof} encodes these fluctuations through a Gaussian constraint in standardized coordinates, constructed below, and Sec.~\ref{subsec:tangent-calibration} describes two ways of sampling $H_0^{\rm comp}$ to calibrate it. 
Both hypotheses are tested against an alternative hypothesis $H_1$ which extends the null family by adding additional components and thus provides the flexibility needed to detect possible deviations.

All expectations under the fitted model are estimated with a reference set $R_0 = \{x_s\}_{s=1}^{S} \sim p_{\hat\theta}$, with $S \gg N_{\rm test}$. 
It is used for the Fisher information, for the normalization of the densities in the test, and to choose the additional components of the alternative hypothesis.

If the model is correctly specified, $\hat\theta$ is asymptotically Gaussian with covariance $F^{-1}/N_{\rm train}$~\cite{redner1984mixture}, where $F$ is the per-event Fisher information. We estimate $F$ on the reference set from the scores of Eq.~\eqref{eq:score},
\begin{equation}
  \widehat F = \frac{1}{S}\sum_{x_s \in R_0} \psi(x_s)\,\psi(x_s)^\top ,
  \label{eq:fisher-hat}
\end{equation}
and take $\widehat{\mathrm{Cov}}(\hat\theta) = \widehat F^{+}/N_{\rm train}$, where $(\cdot)^+$ is a pseudo-inverse with a small ridge.\footnote{Under an incorrect underlying model, the covariance of $\hat\theta$ is instead given by the sandwich formula~\cite{White1982}. We do not use it, since the null distribution is computed assuming the null hypothesis, under which the two estimates coincide. The sandwich then differs from $\widehat F^{+}/N_{\rm train}$ only when the null is false. This choice is also consistent with the parametric bootstrap of Sec.~\ref{subsec:tangent-calibration}, which intrinsically assumes well-specification.} For simplicity of presentation, we write estimates from $R_0$ as averages over the whole dataset, but in practice each estimate uses a subset of $R_0$ for computational efficiency (see App.~\ref{app:implementation}).

We diagonalize $\widehat{\mathrm{Cov}}(\hat\theta) = V\Lambda V^\top$, with eigenvalues $\lambda_1 \ge \dots \ge \lambda_P \ge 0$, and discard the directions with $\lambda_m > \Lambda_{\max}$. These are directions the training data barely constrain~\cite{McLachlanPeel}, along which the fit can move far from $\hat\theta$ and the expansion of Eq.~\eqref{eq:tangent-expansion} breaks down. Collecting the $M \le P$ retained eigenvectors and eigenvalues in $V_M$ and $\Lambda_M$, we define standardized coordinates $v \in \mathbb{R}^M$ through
\begin{equation}
  v = W^{+}\delta\theta, \qquad W \equiv V_M \Lambda_M^{1/2} \in \mathbb{R}^{P\times M},
  \label{eq:whitening}
\end{equation}
which measure each parameter displacement in units of its own uncertainty. 
The resulting coordinates are uncorrelated and have unit variance 
asymptotically, $v \sim \mathcal{N}(0, I_M)$, so that a displacement of 
$\|v\| = 1$ corresponds to a one-sigma shift of the corresponding parameter. 
The corresponding standardized score features,
\begin{equation}
  \Phi(x) \equiv W^\top \psi(x) \in \mathbb{R}^M ,
  \label{eq:whitened-features}
\end{equation}
satisfy
\begin{equation}
  \mathbb{E}_{p_{\hat\theta}}[\Phi] = 0, \qquad
  \mathrm{Cov}_{p_{\hat\theta}}[\Phi] = W^\top \widehat F\, W = \frac{\mathbb{I}_M}{N_{\rm train}},
  \label{eq:unit-variance}
\end{equation}
where the first identity is the vanishing mean of the score and the second follows from $\widehat F = V_M \Lambda_M^{-1} V_M^\top / N_{\rm train}$ on the retained subspace.

\subsection{Uncertainty-aware goodness-of-fit test}
\label{subsec:gof}

We compare data and model with a profile\footnote{Profiling over $v$ is not strictly necessary, but it is desirable because it yields an asymptotically pivotal test statistic (see App.~\ref{app:null-distribution}), improving the behavior of the parametric bootstrap~\cite{hall1992bootstrap}.} likelihood ratio between two exponential families built on $p_{\hat\theta}$.

\paragraph{Null hypothesis.}
To propagate the fit uncertainty into the test, the null hypothesis 
must be promoted from a single fixed density $p_{\hat\theta}$ to a 
family of densities that accounts for the uncertainties; we construct this family by tilting $p_{\hat\theta}$ 
along the uncertainty-normalized score directions. An exponential tilt is used instead of $p_{\hat\theta+Wv}$ because its log-likelihood is linear in $v$, which makes the test a convex problem. The null family reads:
\begin{equation}
  g_v(x) = \frac{p_{\hat\theta}(x)\, e^{v^\top \Phi(x)}}{Z(v,0)} ,
  \label{eq:null-tilt-1}
\end{equation}
with the normalization $Z$ defined in Eq.~\eqref{eq:alternative-2} below. By Eq.~\eqref{eq:tangent-expansion}, we have that $g_v = p_{\hat\theta + Wv}$ to first order in $v$, so the tilt parameters are the standardized parameter displacements of Eq.~\eqref{eq:whitening}, and $H_0^{\rm comp}$ corresponds to $v \sim \mathcal{N}(0,\mathbb{I}_M)$.

\paragraph{Alternative hypothesis.}
\label{sec:tangent-alternative}
The alternative adds additional degrees of freedom via tilts along a fixed dictionary of $J$ localized kernels, following the New Physics Learning Machine (NPLM) approach of Refs.~\citep{DAgnolo:2018cun,NonParametricMethods},
\begin{align}
  h_{v,\alpha}(x) &= \frac{p_{\hat\theta}(x)\, e^{v^\top \Phi(x) + \alpha^\top \tilde k(x)}}{Z(v,\alpha)},
  \label{eq:alternative-1} \\
  Z(v,\alpha) &= \int p_{\hat\theta}(x)\, e^{v^\top \Phi(x) + \alpha^\top \tilde k(x)}\, dx ,
  \label{eq:alternative-2}
\end{align}
so that $g_v = h_{v,0}$. The exponential form keeps $h_{v,\alpha}$ positive for all $(v,\alpha)$. The features $\tilde k_j$ are Gaussian kernels of a common width $\sigma$, $k_j(x) = \exp(-\|x-c_j\|^2/2\sigma^2)$, standardized to zero mean and unit variance on $R_0$, with centers placed by $k$-means clustering of $R_0$ so that they tile the support of the model (see App.~\ref{app:implementation}). The normalization in Eq.~\eqref{eq:alternative-2} is estimated on the reference set,
\begin{equation}
  \widehat Z(v,\alpha) = \frac{1}{S}\sum_{x_s \in R_0} e^{v^\top \Phi(x_s) + \alpha^\top \tilde k(x_s)} .
  \label{eq:zhat}
\end{equation}

The dictionary should tile the support of the model densely enough that no localized departure falls between kernels; we use $J=500$.
Beyond this point additional kernels add little: neighboring kernels of a common width overlap strongly, so the dictionary is highly collinear, and the ridge penalty in Eq.~\eqref{eq:ell} suppresses the redundant combinations.
The effective number of degrees of freedom of the test is therefore set jointly by the kernel width $\sigma$ and the ridge rather than by $J$ itself (see App.~\ref{app:null-distribution}), and our results are stable under moderate changes of $J$.

The width $\sigma$ sets the length scale on which the augmented model Eq.~\eqref{eq:alternative-1} is sensitive to departures from the null family, and its choice is bounded on both sides.
From above, to ensure power, $\sigma$ should not exceed the scale of the discrepancies the test is meant to detect, which for the smooth densities considered here is set by the length scale of the data.
From below, to ensure calibration, $\sigma$ is bounded by the validity of the calibration procedures to be described in Sec.~\ref{subsec:tangent-calibration} (this point is discussed further in App.~\ref{app:resolution}).
To strike a balance between these considerations, we take $\sigma$ to be set by the characteristic scale of the data (alternatively, tests at several widths can be combined \cite{Grosso:2024wjt}); see App.~\ref{app:implementation} for details.

\paragraph{Test statistic.}
On the test sample, the log-likelihood ratio of $h_{v,\alpha}$ to $p_{\hat\theta}$, together with the Gaussian constraint on $v$ and a ridge penalty on $\alpha$, is
\begin{align}
  \ell(v,\alpha) = {} & \sum_{i=1}^{N_{\rm test}} \big[ v^\top \Phi(x_i) + \alpha^\top \tilde k(x_i) \big]
  \nonumber \\-& N_{\rm test} \log \widehat Z(v,\alpha)
   - \tfrac12 \|v\|^2 - \lambda_\alpha \|\alpha\|^2 .
  \label{eq:ell}
\end{align}
Treating the $v$ as nuisance parameters with a Gaussian constraint~\cite{Cowan:2010js}, the test statistic is the profile likelihood ratio
\begin{equation}
  t_{\rm comp} = 2\Big[ \max_{v,\alpha} \ell(v,\alpha) - \max_v \ell(v,0) \Big] .
  \label{eq:tcomp}
\end{equation}
The denominator lets the model move within its uncertainty to accommodate the data, and the numerator additionally switches on the localized distortions. The statistic $t_{\rm comp}$ therefore measures only the incompatibility that the model's uncertainty cannot explain. As an uncertainty-blind baseline, the point-null statistic fixes $v = 0$,
\begin{equation}
  t_{\rm point} = 2 \max_{\alpha} \ell(0,\alpha),
  \label{eq:tpoint}
\end{equation}
with no subtraction needed since $\ell(0,0) = 0$. Because $\ell(v,0)$ is the restriction of $\ell(v,\alpha)$ to $\alpha = 0$, both statistics are nonnegative. For numerical safety, the maximizations over $\alpha$ are also restricted to a box that bounds the log-density distortion of each kernel (see App.~\ref{app:implementation} for details).

Every maximization above is a convex problem. In Eq.~\eqref{eq:ell}, the data term is linear in $(v,\alpha)$, the penalties are concave quadratics, and
\begin{equation}
  \log \widehat Z(v,\alpha) = \log \frac{1}{S}\sum_{x_s\in R_0} e^{v^\top \Phi(x_s) + \alpha^\top \tilde k(x_s)}
  \label{eq:logzhat-lse}
\end{equation}
is a log-sum-exp of linear functions, and hence convex; the box constraint defines a convex set. The objective $\ell$ is therefore strictly concave, and each maximization has a unique optimum, which a quasi-Newton method reaches promptly.

\paragraph{Additional nuisance parameters.} The construction above is the standard profile-likelihood treatment of constrained nuisance parameters~\cite{Cowan:2010js}, with $v$ encoding the statistical uncertainty of the fit. Nothing restricts the nuisance vector to this role: experimental and theoretical systematic uncertainties, parametrized as additional constrained directions in Eq.~\eqref{eq:ell}, can be profiled jointly with $v$, so that a single composite test accounts for all sources of uncertainty on the reference model. This holds provided the systematic variations admit a density parametrization, as in standard template-morphing approaches. We leave this combination to future work and focus here on the fit uncertainty alone, which is the component that existing validation practice omits.

\subsection{Calibration}
\label{subsec:tangent-calibration}

The null distribution of $t_{\rm comp}$ is obtained from $N_B$ pseudo-experiments that sample $H_0^{\rm comp}$. For each, the statistic $t_b$ is computed with the full procedure above, including the same penalties and constraints, and the $p$-value is the rank of the observed statistic,
\begin{equation}
  p = \frac{\#\{t_b \ge t_{\rm obs}\} + 1}{N_B + 1} .
  \label{eq:pvalue}
\end{equation}
We consider two ways of generating the pseudo-experiments. The statistic is the same in both; only its calibration differs.

\paragraph{Gaussian prescription.}
If $\hat\theta$ is Gaussian with the covariance of Sec.~\ref{subsec:uncertainty}, $H_0^{\rm comp}$ can be sampled directly,
\begin{equation}
  z^{(b)} \sim \mathcal{N}(0,\mathbb{I}_M), \quad
  \theta_b = \hat\theta + W z^{(b)}, \quad
  \mathcal{D}_b \sim p_{\theta_b}.
  \label{eq:toys}
\end{equation}
For each pseudo-experiment, a value of $\theta_b$ is chosen, and pseudo-data are then drawn from the GMM at $\theta_b$. 
The test statistic $t_b$ is then computed on this pseudo-data in the same way as it is for the observed data. This procedure suffices to calibrate the observed statistic if the asymptotic approximation is reliable and the Gaussian distribution of $\theta_b$ reproduces the sampling distribution of $\hat{\theta}$.\footnote{The Gaussian prescription is the frequentist analogue of Laplace resampling, differing in that it builds a frequentist null distribution rather than a Bayesian posterior predictive distribution.}
\begin{figure*}[htb]
    \centering
    \includegraphics[width=0.8\linewidth]{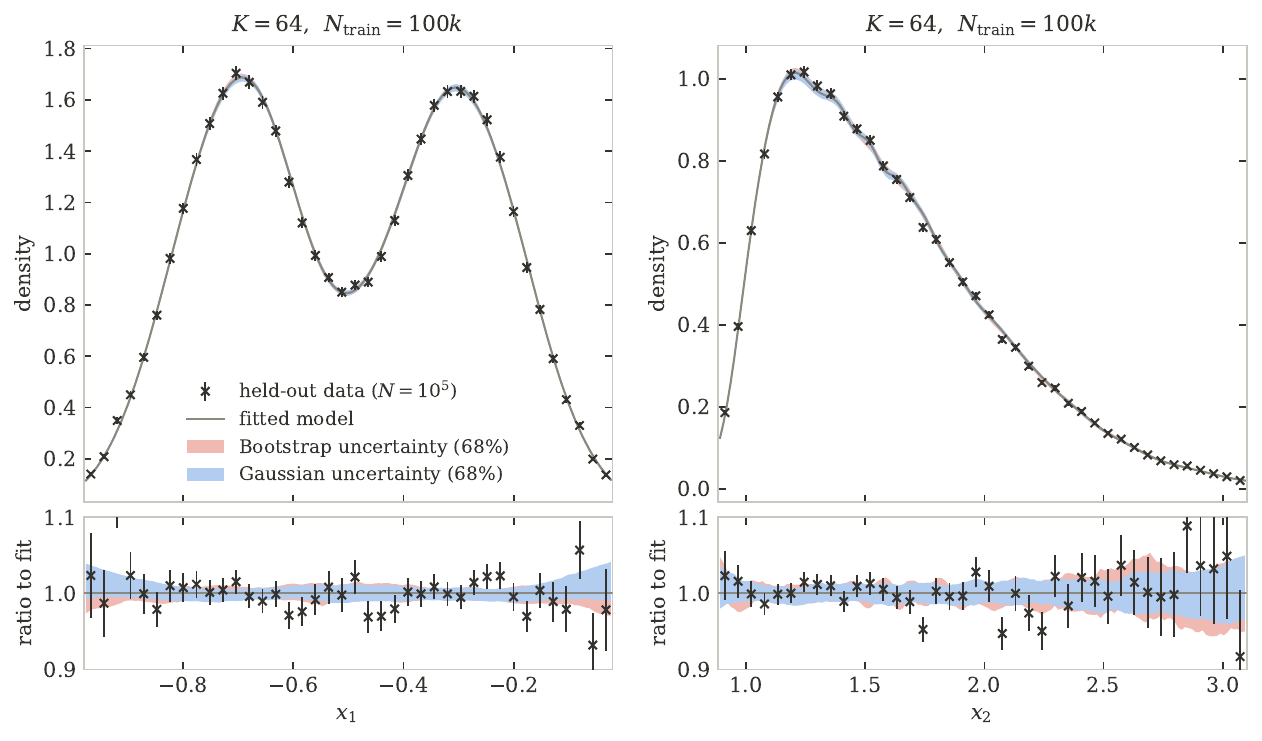}
    \caption{\textbf{The marginal distribution of the two features in the 2D toy problem}. The black crosses show the binned density evaluated with $10^5$ held-out events. The solid line shows the marginals of a GMM with $K=64$ components fit to $N_\text{train}=100,000$ events, and the colorful envelopes are uncertainties corresponding to the calibration prescriptions of Sec.~\ref{subsec:tangent-calibration}. Specifically, the blue envelope corresponds to the propagated Gaussian uncertainties on the best-fit parameters, and the red envelope is obtained as the $0.16$ and $0.84$ quantiles of the refit models obtained using a parametric bootstrap.}
    \label{fig:2d_marginals}
\end{figure*}

\paragraph{Bootstrap prescription.}

For a GMM with many parameters and a multimodal likelihood, the Gaussian description of $\hat\theta$ can fail: expectation maximization~(EM) may converge to different local maxima for different training samples~\cite{redner1984mixture}, and the scatter of the fit need not be Gaussian, nor small enough for Eq.~\eqref{eq:tangent-expansion} to hold.

The parametric bootstrap relaxes this assumption by directly simulating the fitting procedure. Treating $p_{\hat\theta}$ as if it were the true density, we repeatedly draw training samples from it and refit; the spread of the refits $\hat\theta_b$ around $\hat\theta$ then serves as a proxy for the (unobservable) spread of $\hat\theta$ around the truth. The necessary assumption is that the spread from the fitting procedure is similar whether the data come from $p_0$ or from the nearby $p_{\hat\theta}$, which is weaker than the asymptotic conditions required for the Gaussian prescription. However, this assumption can fail even for a well-specified model, and in particular we find in App.~\ref{app:resolution} that the bootstrap calibration breaks down if the kernel width of Sec.~\ref{subsec:gof} is chosen too narrow compared to the characteristic scale of the data.

In practice, for each pseudo-experiment, we draw a training sample of size $N_{\rm train}$ from $p_{\hat\theta}$ and refit the full GMM with EM from a fresh $k$-means initialization, obtaining $\hat\theta_b$. We rebuild the entire construction at $\hat\theta_b$ exactly as for the nominal fit (i.e. the reference set, the Fisher information, the standardization, and the kernel dictionary), draw a test sample of size $N_{\rm test}$ from $p_{\hat\theta}$, and compute $t_b$.

These two prescriptions for generating the calibration distribution are, in a sense, mirror images of each other. The Gaussian prescription holds the fit fixed at $\hat{\theta}$ and moves the data-generating model to $\theta_b$, while the bootstrap prescription moves the fit to $\hat{\theta}_b$ and holds the data-generating model fixed at $\hat{\theta}$. Both, then, should reproduce the displacement between fit and truth, but the bootstrap does not assume that this displacement is small or Gaussian, and it captures the optimization variability. We will find that it remains calibrated to larger $N_{\rm test}$ (see Sec.~\ref{sec:exp}) and is our nominal prescription. The Gaussian prescription requires no refits and provides an explicit parametrization of the model uncertainty that can be easily propagated to the density and to downstream applications, and it remains useful where its assumptions hold.
The two prescriptions also differ in their exposure to the truncation choice: $\Lambda_{\max}$ enters the Gaussian prescription twice, through the profiled directions in the test and through the fluctuations injected in the pseudo-experiments, whereas the bootstrap refits are constructed independently of it and dependence appears only in the test. As such, scanning $\Lambda_\text{max}$, we find in App.~\ref{app:implementation} that the bootstrap calibration is robust against the choice of $\Lambda_{\max}$, while the Gaussian prescription shows a stronger dependence.

In terms of the sources of Sec.~\ref{sec:motivation}, the Gaussian prescription captures the data variability around a single optimum, while the bootstrap, which refits the model from a fresh initialization, also captures the optimization variability. This is why, in practice, we find that the bootstrap prescription yields uncertainty 
bands that are comparable to or larger than those of the Gaussian prescription. Neither, however, accounts for the model bias. 

For the point-null baseline, the pseudo-experiments are simply drawn from $p_{\hat\theta}$ with the construction held fixed.

\paragraph{Asymptotic $\chi^2$ limit.}
In the quadratic approximation to $\ell$ and without the ridge penalty $\lambda_\alpha$ on the $\alpha$, $t_{\rm comp}$ follows a $\chi^2_J$ distribution under the Gaussian composite-null, as shown in App.~\ref{app:null-distribution}. The same holds for $t_{\rm point}$ under the point-null, and the ridge lowers the effective number of degrees of freedom below $J$. We will see in Sec.~\ref{sec:exp} that this approximate $\chi^2_J$ prediction holds (up to a decrease in the effective degrees of freedom due to $\lambda_\alpha$) in the regime where the asymptotics are reliable and the Gaussian and bootstrap prescriptions are in good agreement, but that it breaks down outside of this regime (as it should, since the derivation of this $\chi^2_J$ distribution assumes Gaussianity). As such, our results for test calibration are always evaluated using pseudo-experiments.\\

It is worth noting that the test described in Sections~\ref{subsec:gof} and~\ref{subsec:tangent-calibration} is independent of how the density is estimated and of how its uncertainty is obtained, and it works provided this uncertainty is reliably estimated and can be propagated into the composite test.
\begin{figure*}[htb]
    \centering
    \includegraphics[width=1.0\linewidth]{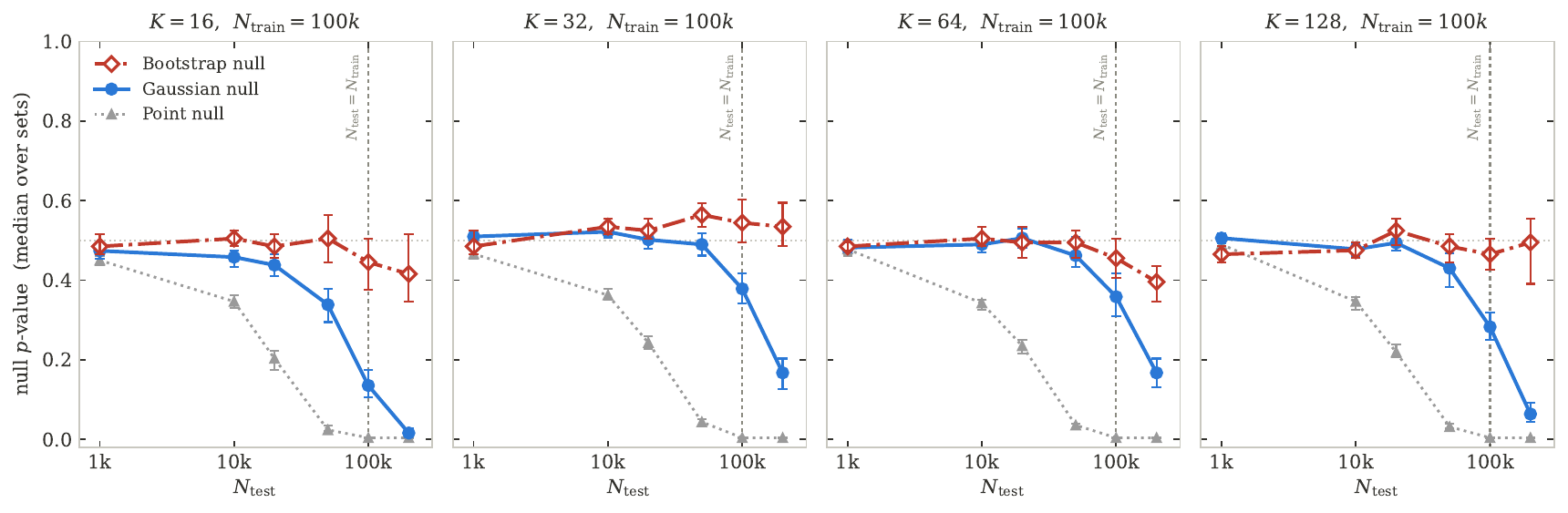}
\caption{\textbf{Closure test for the 2D toy problem.} Median null $p$-value for 
the three goodness-of-fit tests of Sec.~\ref{subsec:gof} versus test-set 
size $N_\mathrm{test}$, for $K = 16, 32, 64, 128$ (left to right) at fixed 
$N_\mathrm{train} = 10^5$. A calibrated test gives median $p = 0.5$ 
(horizontal dotted line); the vertical dashed line marks 
$N_\mathrm{test} = N_\mathrm{train}$. Error bars are 68\% confidence 
intervals on the median.}
    \label{fig:closure-2d-100k}
\end{figure*}

\section{Numerical experiments}\label{sec:exp}
In this section we apply the method presented in Sec.~\ref{sec:method} to two problems: a two-dimensional toy model with known analytic ground truth and a four-dimensional problem from particle physics.

\subsection{Toy model}
\label{subsec:toy}
We first validate the pipeline on a controlled two-dimensional toy problem, where the ground truth is known analytically and can be resampled at will.
The target factorizes into two independent features, each chosen to stress a different aspect of the fit:
\begin{itemize}
    \item \textbf{Feature 1}: a bimodal Gaussian mixture, $p(x_1) = \tfrac{1}{2}\mathcal{N}(x_1;\,-0.70,\,0.12^2) + \tfrac{1}{2}\mathcal{N}(x_1;\,-0.30,\,0.12^2)$, whose modes are separated by more than three times their common width, with analytic mean $\mu_1^\star = -0.50$.
    \item \textbf{Feature 2}: a skew-normal distribution with location $\xi=1.0$, scale $\omega=0.75$ and shape $\alpha=8.0$, i.e.\ strong positive skew, with analytic mean $\mu_2^\star = \xi + \omega\,\delta\sqrt{2/\pi}$ and $\delta = \alpha/\sqrt{1+\alpha^2}$.
\end{itemize}
The first feature probes the ability of the model to resolve well-separated modes, the second its ability to reproduce a long, asymmetric tail.

We draw $N_{\rm train}=10^{5}$ training points and fit the Gaussian mixture model of Eqs.~\eqref{eq:gmm} and~\eqref{eq:gmm-param} for four choices of the number of components, $K=16,\,32,\,64,\,128$.
We then estimate the model uncertainties following Sec.~\ref{subsec:uncertainty} and assess the quality of the fit with the point-null test of Eq.~\eqref{eq:tpoint}, which treats the fitted density as exact, and the composite test of Eq.~\eqref{eq:tcomp}, which accounts for the uncertainty of the fit.
For the composite test we report both our default bootstrap calibration as well as the Gaussian prescription of Sec.~\ref{subsec:tangent-calibration}.
Because the ground truth is available in closed form, a fresh test sample is drawn for every replica, so the closure test is not subject to reuse of validation data across toys.

Figure~\ref{fig:2d_marginals} shows the marginal distributions of the two features together with a $K=64$ fit.
The fitted model (black solid line) is compared with the empirical distribution of a held-out sample the same size as the training set ($\times$ markers), with the data-to-fit ratio shown in the bottom panels.
The 68\% bands are built under the Gaussian prescription (light blue) and under the bootstrap prescription (light red).
The Gaussian and bootstrap bands are overall comparable, with the bootstrap band slightly larger in regions where the
Gaussian approximation underestimates the spread of the fit around the truth. 

Figure~\ref{fig:closure-2d-100k} shows the median $p$-value for the point null as well as the Gaussian and bootstrap composite nulls as a function of the test-set size $N_{\rm test}$ for the four values of $K$. 
For each curve, a median $p$-value of $0.5$ would indicate correct calibration, since under the null hypothesis the $p$-value is uniformly distributed by construction.
We perform 20 independent model fits for each value of $K$, each evaluated on 100 test datasets per value of $N_\mathrm{test}$. 
The point-null and Gaussian composite statistics are calibrated with 250 pseudo-experiments per fit, while the bootstrap calibration uses 100 refits in total per fit, with the same refits reused across all values of $N_\mathrm{test}$. 
The point-null test is calibrated only in the small-sample regime. 
Its median $p$-value is compatible with $0.5$ for $N_{\rm test} < N_{\rm train}/10$, starts degrading beyond that point, and collapses to zero by $N_{\rm test}=N_{\rm train}$ for every value of $K$ we consider. 
The behavior is essentially independent of the number of mixture components $K$, as the test fails because it neglects the uncertainty of the fit, regardless of the flexibility of the model.

Promoting the test to the composite-null, with the model uncertainties estimated under the Gaussian prescription, largely restores calibration at intermediate values of $N_\text{test}$.
The median $p$-value stays compatible with $0.5$ up to about $N_\text{test} = N_\text{train}/5$, but calibration still fails dramatically at $N_\text{test} = 2 N_\text{train}$, with a median $p$-value less than $0.2$ for all $K$.

The bootstrap calibration prescription fully restores closure, remaining calibrated up to $N_\text{test} = 2 N_\text{train}$ for all $K$. 
This is consistent with the bootstrap capturing the optimization variability and the non-Gaussian scatter of the fit, which the Gaussian description misses.

\begin{figure}[h]
    \centering
    \includegraphics[width=0.48\textwidth]{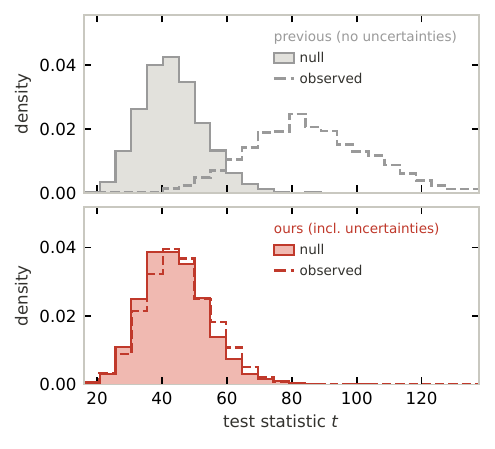}
    \caption{\textbf{Calibrated uncertainties make the goodness-of-fit test cover.}
    Distribution of the test statistic $t$ for a GMM fit with $K=64$ components at $N_{\rm train}=N_{\rm test}=10^{5}$ to the 2D toy model.
    The observed point statistic of Eq.~\eqref{eq:tpoint} (dashed gray step histogram) lies almost entirely above its point-null (gray filled histogram), which treats the fitted density as exact, so the point-null test rejects the ground truth essentially always.
    The observed composite statistic of Eq.~\eqref{eq:tcomp} (dashed red step histogram) is instead reproduced by its bootstrap null (red filled histogram), which propagates the uncertainty of the fit, so the composite test closes. Each distribution is estimated using $20$ model fits, each of which is used to evaluate the point and composite test statistic on $100$ test datasets for each fit. The point-null is then estimated with $250$ point-null toys and the bootstrap null is estimated with $100$ bootstrap toys for each fit.}
    \label{fig:money-plot-2d}
\end{figure}

These results demonstrate that the point-null hypothesis is insufficient. 
We find that closure is dramatically improved through the use of a composite-null hypothesis, and that maintaining closure for $N_\text{test} \gtrsim N_\text{train}$ requires bootstrap (rather than Gaussian) calibration. 
Figure~\ref{fig:money-plot-2d} illustrates this directly for $K=64$ at $N_\mathrm{test} = N_\mathrm{train}$: the observed point-null test statistic lies almost entirely above its null distribution, while the composite-null statistic is well reproduced by its bootstrap null.

\subsection{Embedded particle jets}
\label{subsec:jets}

Recent advances in representation learning have made it possible to construct powerful neural embeddings of highly complex physics data. 
Among other applications, these embeddings have enabled anomaly detection in a latent space~\cite{Dillon:2022tmm,Harris:2024sra,Matos:2024ggs,Metzger:2025ecl,bright2025autoscidact,Li:2026azw}.
Anomaly detection in a latent space, however, requires a robust understanding of the density of the reference model. 
The method proposed in this paper offers a path to estimate a background data distribution and its uncertainties with statistical guarantees. 

As a demonstration, we emulate a realistic physics analysis aiming to identify anomalous jet decays. 
We use the \textsc{JetClass} dataset~\cite{Qu:2022mxj}, a large-scale public benchmark for jet tagging. 
The \textsc{JetClass} dataset provides 100M simulated jets for training, 5M for validation, and 20M for testing. 
Each jet is represented as a particle cloud of charged and neutral hadrons, electrons, muons, and photons and labeled by the initiating parton and decay process, with ten equally represented classes. 
We restrict the analysis to purely hadronic final states and consider five classes: decay of top quarks, W and Z bosons, Higgs bosons and QCD multijet production.

\begin{figure}[!ht]
    \centering
    \includegraphics[width=1.0\columnwidth]{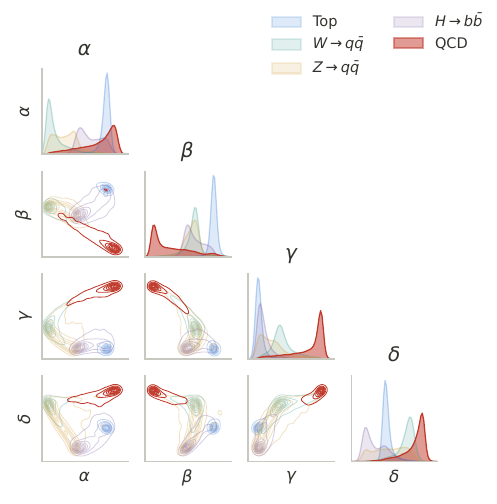}
    \caption{\textbf{Particle jets embedded in a 4D space.} Corner plot of the four-dimensional jet embedding learned with a supervised contrastive (SupCon) objective~\cite{Khosla:2020supcon}. Diagonal panels show the one-dimensional marginal distributions of each embedding coordinate $(\alpha, \beta, \gamma, \delta)$; off-diagonal panels show the pairwise joint densities as contours. The QCD multijet density (red) is highlighted because it is the density we apply our framework to; other classes are used to obtain the embedding and are shown for reference only.}
    \label{fig:embedding}
\end{figure}

\begin{figure*}[!ht]
    \centering
    \includegraphics[width=0.8\linewidth]{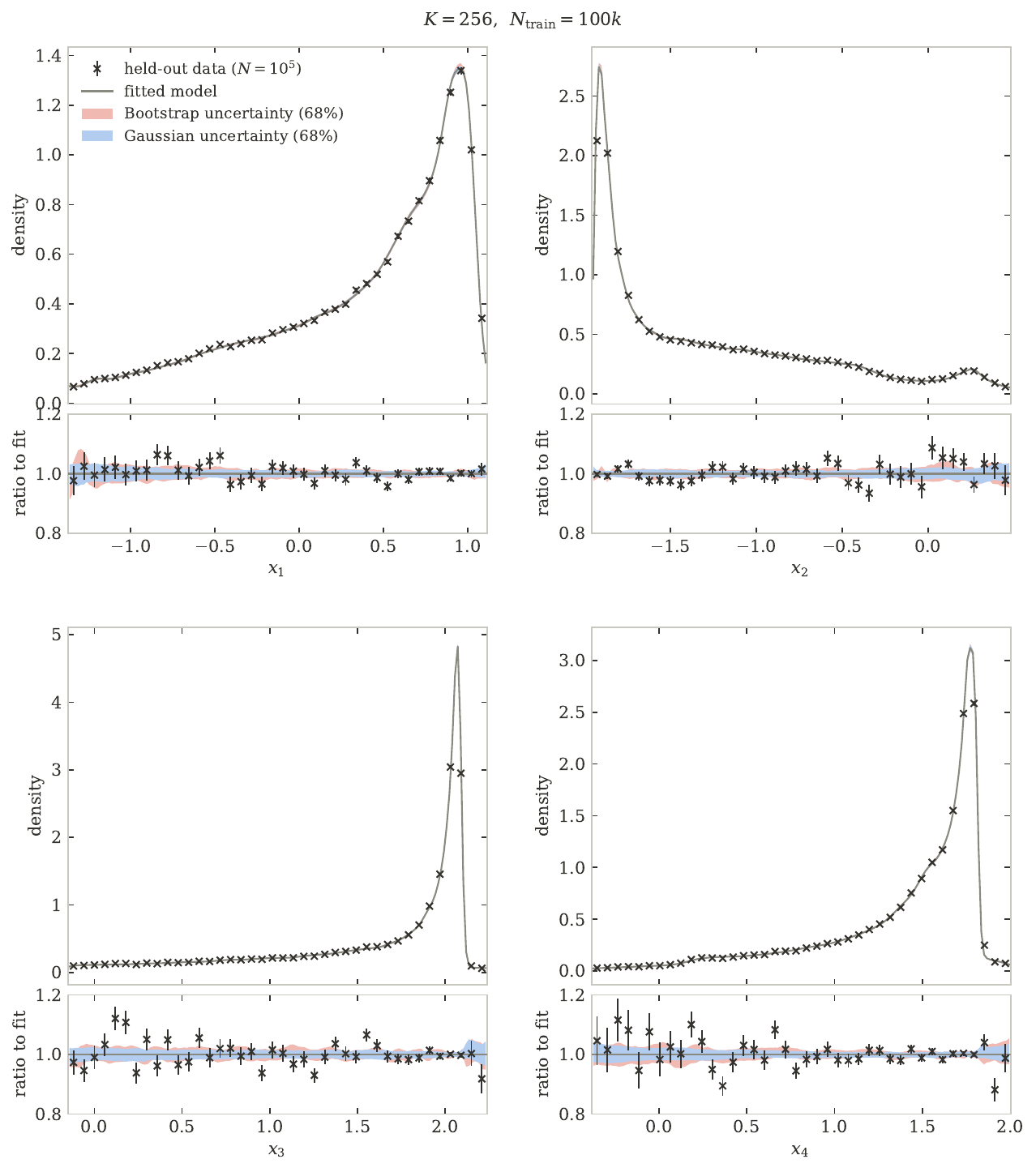}
    \caption{\textbf{The marginal distribution of the four features in the 4D particle jets problem.} The black crosses show the binned density evaluated with $10^5$ held-out events. The solid line shows the marginals of a GMM with $K=256$ components fit to $N_\text{train}=100,000$ events, and the colorful envelopes are uncertainties corresponding to the calibration prescriptions of Sec.~\ref{subsec:tangent-calibration}. Specifically, the blue envelope corresponds to the propagated Gaussian uncertainties on the best-fit parameters, and the red envelope is obtained as the $0.16$ and $0.84$ quantiles of the refit models obtained using a parametric bootstrap.}
    \label{fig:4d_marginals}
\end{figure*}
\begin{figure*}[!ht]
    \centering
    \includegraphics[width=1.0\linewidth]{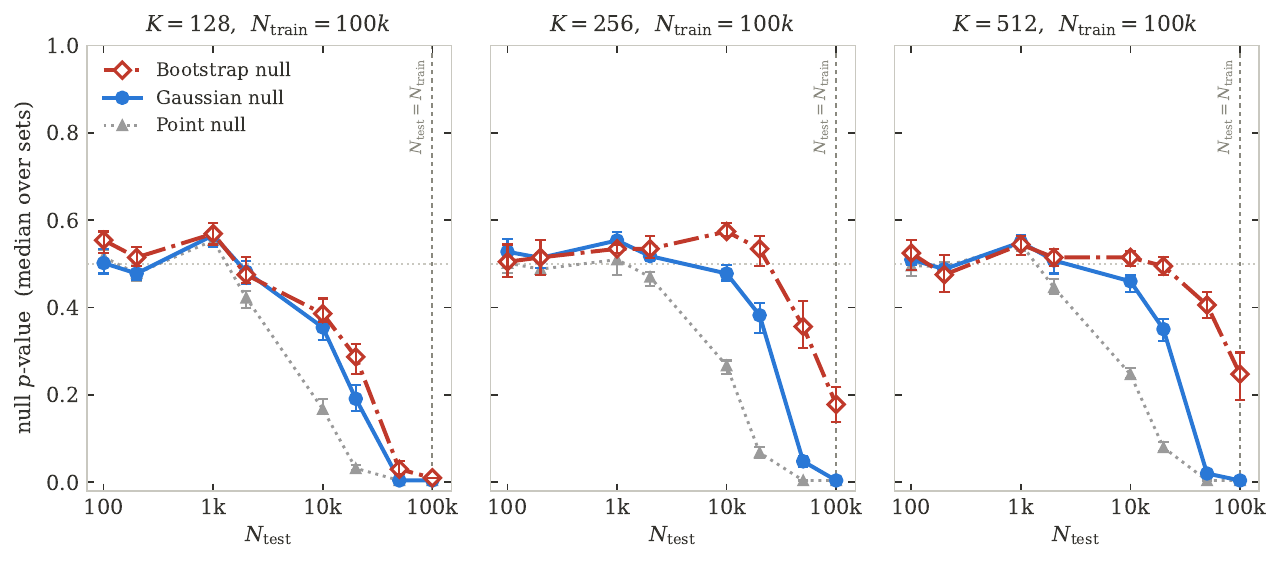}
    \caption{\textbf{Closure test for the 4D particle jets problem.} Median null $p$-value for the three goodness-of-fit tests of Sec.~\ref{subsec:gof} versus test-set size $N_\mathrm{test}$, for $K = 128, 256, 512$ (left to right) at fixed $N_\mathrm{train} = 10^5$. A calibrated test gives median $p = 0.5$ (horizontal dotted line); the vertical dashed line marks $N_\mathrm{test} = N_\mathrm{train}$. Error bars are 68\% confidence intervals on the median.}
    \label{fig:closure-4d-100k}
\end{figure*}

%
Following Ref.~\cite{bright2025autoscidact}, the embedding is trained with a SupCon (contrastive) objective~\cite{Khosla:2020supcon}, which extends self-supervised contrastive learning~\cite{Chen:2020simclr} to labeled data: all jets sharing a class label are treated as positives, and jets from different classes as negatives. 
For training the embedding, we restrict the \textsc{JetClass} training and validation set to 10\% of the available statistics, corresponding to 5M and 250k jets, respectively, equally distributed among the five considered classes.
The resulting representation is shown in Fig.~\ref{fig:embedding}.

Anomaly detection via two-sample testing~\cite{DAgnolo:2018cun,LearningMultivariateNP,NeymanPearsonGoF} requires either a large reference (simulated) sample or an equivalent of the underlying probability density function in the embedding space.
Since large simulated samples can be prohibitively expensive to generate, we apply the \methodname method as described in Sec.~\ref{sec:method} to model the QCD background in the embedding space.\footnote{We do not claim that using \methodname allows us to extract information that was not already in the simulated sample. Rather, background modeling with \methodname is motivated because its quantified uncertainties and composite null are necessary to avoid false positives from finite-simulation-sample error on the background estimate.} 
We choose QCD multijet production because it is a dominant background whose simulated statistics are hard to match to the large number of events expected in LHC data. 
We apply our pipeline to the full \textsc{JetClass} QCD multijet test partition with $N_{\rm test}^{\rm TOT}= 2\text{M}$ jets, which we split into a training pool and a testing pool each consisting of 1M jets. The dataset can be found on Zenodo~\cite{reissel_2026_23693193}. 

Each \methodname training and $p$-value evaluation is performed by sampling from these finite pools, so we report closure results only up to $N_\text{test} = N_\text{train} = 10^5$ events to mitigate correlations between different trainings and draws of pseudo-observed data.

Figure~\ref{fig:4d_marginals} shows the marginal distributions of the four latent features of our embedding for the QCD sample, together with a $K=256$ fit. As for the two-dimensional problem, we compare the fitted model (black solid line) with the empirical distribution of a same-size held-out sample ($N=10^5$) marked by the ``$\times$'' symbol. The data-to-fit ratio is shown in the bottom panels.
The 68\% bands are built under the Gaussian prescription (light blue) and under the bootstrap prescription (light red).

Figure~\ref{fig:closure-4d-100k} shows the median $p$-value for the goodness-of-fit test using the point-null, the Gaussian composite and the bootstrap composite as a function of $N_{\rm test}$ for $K=128, \, 256,\, 512$. 
For each value of $K$, we perform 10 independent model fits, each evaluated on 100 test datasets per value of $N_\mathrm{test}$. 
The point-null and Gaussian composite statistics are calibrated with 250 pseudo-experiments per fit; since the bootstrap calibration is considerably more expensive, it uses 100 refits in total per fit, with the same refits reused across all values of $N_\mathrm{test}$.

The results are qualitatively similar to those in 2D.
At all values of $K$, the point-null test starts degrading before $N_{\rm test}\sim N_{\rm train}/10$.
The composite test with the Gaussian prescription improves the calibration, enabling closure up to $N_{\rm test}\sim N_{\rm train}/10$ for $K\ge 256$.
The bootstrap prescription further improves the results allowing closure up to $N_{\rm test}=N_{\rm train}/5$ for $K \geq 256$, with calibration degrading noticeably (but not completely) for larger $N_\text{test}$.

The $K=128$ \methodname model performs visibly worse than either of the two larger models, and the difference between the Gaussian and bootstrap null prescriptions is substantially smaller than for larger $K$.
We interpret this as underparametrization-induced model bias in the sense of Sec.~\ref{sec:motivation}, where a mixture model with only $128$ components is insufficiently expressive to fit the data.
This model misspecification is not quantified in our statistical uncertainties, so sufficiently large $N_\text{test}$ necessarily rejects this model.
By contrast, the calibration is evidently saturating with $K$ for large $N_\text{test}$ for the $K=256,512$ models, with the larger difference between the bootstrap and Gaussian prescriptions indicating a breakdown in the asymptotics.
This suggests that the finite $N_\text{train} = 10^5$, rather than model capacity, is likely the primary barrier to achieving a more performant \methodname in the context of this case study, as could be expected from fitting a model with several thousand parameters.
However, even with these limitations, the bootstrap composite null is calibrated to values of $N_\text{test}$ an order of magnitude larger than the point-null.
To characterize the extent to which these results are specific to \methodname models, we also test parametrized ensembles of normalizing flows,\footnote{We only consider flows with the point-null and Gaussian composite-null prescriptions, as the bootstrap-null would require retraining the whole ensemble of flows in each pseudo-experiment, which would be prohibitively expensive.} and the results can be found in App.~\ref{app:nf-results}.
In short, we find that the \methodname models achieve substantially better calibration properties than even ensembles of $\mathcal{O}(10^2)$ flows, while being dramatically cheaper to train and evaluate.

\section{Conclusion}\label{sec:conclusion}
All generative models trained on finite samples are inexact approximations of their training distribution, and a sufficiently powerful goodness-of-fit test will eventually detect deviations. This is a consequence of finite information, not a failure of design. A density estimated from $N_{\rm train}$ samples carries an inherent estimation error, so even a perfectly specified model is rejected by a powerful goodness-of-fit test. 
Validating a generative model against a point-null therefore measures the size of the test sample rather than the quality of the model, and no architectural or training improvement can change this.

An approximate model can nonetheless be useful, provided its estimation error is known.
In this work, we make this statement quantitative. By linearizing a Gaussian mixture model around its best fit, we obtain the covariance of its parameters from its analytic score functions and Fisher information, and the goodness-of-fit test becomes a convex problem. We find that promoting the null from a point hypothesis to a composite hypothesis, together with a parametric bootstrap calibration procedure, is able to calibrate the goodness-of-fit test on datasets an order of magnitude larger than the point-null equivalent.

The success of \methodname is in contrast with neural generative models trained through stochastic gradient descent, which typically do not converge to a well-defined optimum and which have non-invertible information matrices due to parameter degeneracies. We find in App.~\ref{app:nf} that even a parametrized ensemble of normalizing flows is only able to capture part of the uncertainty, and cannot match the performance of \methodname in our low-dimensional case studies.

Uncertainty quantification for generative models bears directly on how generated samples can be used. 
A generative model cannot create information that was not in its training data~\citep{Prasanth}. 
What it can do is produce arbitrarily many samples whose deviation from the truth is bounded by stated uncertainties. 
Generated samples can therefore be used beyond the training statistics only if these uncertainties are carried along into the downstream analysis, e.g. by using a composite-null hypothesis in the context of a goodness-of-fit or hypothesis test.
For statistical inference, the true figure of merit of a generative surrogate is therefore the precision to which it can be trusted.

Within this framework, \methodname still has clear limitations. 
Because the number of parameters of a GMM grows quickly with dimension, the method is suited to low-dimensional spaces rather than directly to high-dimensional ones. 
This further motivates the compression of high-dimensional data into a small number of summary statistics, such as through learned embeddings~\cite{Dillon:2022tmm,Harris:2024sra,Matos:2024ggs,Metzger:2025ecl,bright2025autoscidact,Li:2026azw}.
Furthermore, \methodname accounts for the statistical variability of the fit due to the finite training sample and to the stochasticity of optimization, but it is still subject to model bias: a model family that cannot represent the data will still be rejected by a sufficiently powerful test.
The need to reduce model bias makes the choice of $K$ a trade-off. Increasing the number of components reduces mismodeling but it also adds parameters, leading to larger uncertainties and potentially undermining the calibration procedures.
The Gaussian calibration manifestly requires the asymptotic regime where $N_\text{train} \gg P$, and the bootstrap calibration relies on this indirectly through the assumption that the estimator spread is similar for fits to draws from $p_0$ and from $p_{\hat{\theta}}$. 

We currently choose $K$, as well as the truncation hyperparameter $\Lambda_{\max}$, empirically. These model selection choices, as well as verifying that the model can be trusted beyond its training statistics, require closure tests on held-out samples at least as large as the desired inference dataset. 
This is costly in exactly the simulation-scarce settings where generative models are most useful, and the necessity of the bootstrap for the most trustworthy calibration adds the cost of refitting the model for every pseudo-experiment.

Each of these limitations offers promising avenues for progress.
Both the restriction to low-dimensionality and the model bias considerations motivate exploring distinct model classes.
In particular, the test of Sec.~\ref{subsec:gof} is agnostic to the model class: it requires only tractable score vectors, which automatic differentiation supplies for any neural density estimator, and a faithful estimate of the uncertainty of the fit. 
Our tangent-space construction is then the exact, low-dimensional counterpart of the linearized Laplace approximations developed for deep networks~\cite{immer2021improving, daxberger2021laplace}, and that literature's low-rank and subspace curvature estimates offer a natural route to extending the composite-null test to larger models; the open problem is promoting such uncertainties from predictive heuristics to calibrated ingredients of a goodness-of-fit test.
On the other hand, the need for empirical tuning of $K$ and $\Lambda_\text{max}$ could be eliminated with a data-driven procedure for hyperparameter selection.
Such a procedure could involve the use of cross-validation to check closure on subsets of held-out simulation data.
Moreover, it would be interesting to consider applications with systematic uncertainties.
Because the fit uncertainty enters as a standard constrained nuisance vector, the Gaussian calibration prescription composes naturally with the existing treatment of systematic uncertainties~\cite{dAgnolo:2021aun, Tong:2026}: additional nuisance directions are profiled jointly, and a single composite test accounts for all sources of uncertainty on the reference model.
In contrast, composing our nominal bootstrap calibration prescription (which does not yield straightforwardly parametrized uncertainties) with additional nuisance parameters may be more challenging.

The immediate application of our method is background estimation for anomaly detection in low-dimensional learned embeddings. 
There, a validated model could enter the final test in closed form, together with its uncertainties, as the composite-null hypothesis. 
More broadly, as generative models increasingly substitute for simulation across various scientific fields, uncertainty-aware validation is necessary for these surrogates to serve as instruments of trustworthy statistical inference.

The code required to reproduce our experiments and implement \methodname can be found here: \url{https://github.com/gbadarac/UQ-GoF-GenModels/tree/paper}.

\section*{Acknowledgments}
We are especially grateful to Thea {\AA}rrestad for her supervision, guidance and support in the development of this work. We also thank Davide Valsecchi for discussions on uncertainty quantification for generative models and for his detailed feedback on the manuscript. We further thank Prasanth Shyamsundar for conversations on the information content of generated samples and the limits of data amplification.
The authors acknowledge the financial support of the National Science Foundation under Cooperative Agreement PHY-2019786 (NSF AI Institute for Artificial Intelligence and Fundamental Interactions, http://iaifi.org/), and the National Science Foundation AI Expand program~(OAC-2334265). C.R. further acknowledges support under a Postdoc.Mobility
fellowship from the Swiss National Science Foundation (SNSF) (Project No. P500PT 222340). Claude (Anthropic) assisted with manuscript editing and code development. All scientific ideas and experiments are the original work of the authors. 

\appendix
\section{Additional discussion on the limits of generative models}
\label{app:KL_discussion}

In this appendix, we quantify the statistical uncertainty of the fit introduced in Sec.~\ref{sec:motivation} in terms of the Kullback--Leibler (KL) divergence between the true and the fitted densities, and derive the test sample size at which it becomes detectable by a goodness-of-fit test.
We consider a parametric family $p_\theta$ with $\theta\in\mathbb{R}^{P}$ and assume the best-case scenario of Sec.~\ref{sec:motivation}: the model is well specified, $p_{0}=p_{\theta_{0}}$ for some interior point $\theta_{0}$ with finite, non-singular Fisher information $F(\theta_{0})$, and the optimizer returns the global maximum-likelihood estimate $\hat\theta$ computed on $N_{\text{train}}$ independent samples of $p_{0}$ (i.e.\ we isolate the data variability; any optimization variability would only add to the result below).

Expanding $\theta\mapsto\mathrm{KL}(p_{0}\,\|\,p_{\theta})$ around $\theta_{0}$, the zeroth-order term vanishes because $p_{0}=p_{\theta_{0}}$, and the gradient vanishes because $\theta_{0}$ minimizes the divergence over the family. The curvature at the minimum is the Fisher information: since the $\theta$-dependence of the divergence sits entirely in the cross-entropy $-\mathbb{E}_{p_{0}}[\log p_{\theta}]$, its Hessian is 
\begin{equation}
    F(\theta_{0}) = -\mathbb{E}_{p_{0}}\big[\nabla_\theta\nabla_\theta^{\top}\log p_{\theta}\big]_{\theta_{0}}.
\end{equation} 
To leading order, therefore,
\begin{equation}
  \mathrm{KL}(p_{0}\,\|\,p_{\hat\theta})
  \;\simeq\; \tfrac{1}{2}\,(\hat\theta-\theta_{0})^{\top} F(\theta_{0})\,(\hat\theta-\theta_{0})\,:
  \label{eq:kl-quadratic}
\end{equation}
the divergence of the fitted density is a quadratic form in the estimation error, with the information matrix as its metric.
The maximum-likelihood estimate asymptotically fluctuates around the true value as $\hat\theta - \theta_{0} \sim \mathcal{N}\!\big(0,\, F(\theta_{0})^{-1}/N_{\text{train}}\big)$, and for a Gaussian vector $u\sim\mathcal{N}(0,\Sigma)$ one has $\mathbb{E}[u^{\top}\!Au]=\mathrm{tr}(A\Sigma)$. Averaging Eq.~\eqref{eq:kl-quadratic} over the training sample then gives, for large enough $N_{\text{train}}$,
\begin{equation}
  \mathbb{E}\big[\,\mathrm{KL}(p_{0}\,\|\,p_{\hat\theta})\,\big]
  \;\simeq\; \frac{\mathrm{tr}\big(F F^{-1}\big)}{2\,N_{\text{train}}}
  \;=\; \frac{P}{2\,N_{\text{train}}}\,.
  \label{eq:kl-scaling}
\end{equation}
This is the data variability, expressed in terms of the KL divergence.
Note that the Fisher information cancels against its own inverse: every identifiable direction in parameter space contributes $1/(2N_{\text{train}})$ nats to the expected divergence, independently of how tightly the data constrain it. A poorly measured direction fluctuates more, but the divergence is correspondingly flat along it, and the two effects compensate. Only the number of directions, $P$, survives. Equivalently, $2N_{\text{train}}\,\mathrm{KL}(p_{0}\,\|\,p_{\hat\theta})$ is asymptotically the Wald statistic $N_{\text{train}}(\hat\theta-\theta_0)^\top F(\hat\theta-\theta_0)$, which, like the likelihood-ratio statistic of Wilks' theorem, is $\chi^{2}_{P}$-distributed with mean $P$. For the Gaussian example of Fig.~\ref{fig:pvalue-degradation}, $P=2$, and indeed $\mathrm{KL}\simeq\hat\mu^2/2+(\hat\sigma-1)^2$ with $\mathrm{Var}(\hat\mu)=1/N_{\text{train}}$ and $\mathrm{Var}(\hat\sigma)=1/(2N_{\text{train}})$, so that $\mathbb{E}[\mathrm{KL}]\simeq 1/N_{\text{train}}$.

A goodness-of-fit test performed with $N_{\text{test}}$ samples becomes sensitive to a discrepancy of size $D$ (in KL) once $N_{\text{test}}\,D \gtrsim \mathcal{O}(1)$. This is the Neyman--Pearson benchmark: for the most powerful test of $p_{\hat\theta}$ against the specific alternative $p_{0}$, the expected log-likelihood ratio per test sample is $\mathbb{E}_{p_{0}}[\log(p_{0}/p_{\hat\theta})]=\mathrm{KL}(p_{0}\,\|\,p_{\hat\theta})=D$, so the evidence accumulated over the test set, $N_{\text{test}}\,D$, must exceed its $\mathcal{O}(1)$ null fluctuations. Substituting Eq.~\eqref{eq:kl-scaling}, which represents the discrepancy that the most powerful test would detect, this threshold is crossed once
\begin{equation}
  N_{\text{test}} \;\gtrsim\; \frac{N_{\text{train}}}{P}\,.
  \label{eq:ntest-scaling}
\end{equation}
This argument should be understood as establishing a sensitivity ceiling for goodness-of-fit tests, describing when the training-set discrepancy becomes detectable in principle, not necessarily when a specific test statistic used in practice (such as the test statistics described in Sec.~\ref{subsec:gof}) rejects.
The actual rejection point of a test depends on its power against this particular alternative and on its empirical calibration, and can in general occur later, and with a different $\mathcal{O}(1)$ prefactor, than this bound suggests: a statistic built on $J$ degrees of freedom, for instance, resolves the same $D$ only once $N_{\text{test}}\,D\gtrsim\sqrt{J/2}$, which rescales Eq.~\eqref{eq:ntest-scaling} by a constant without affecting the scaling in $N_{\text{train}}$ and $P$.

Two qualifications delimit the validity of Eq.~\eqref{eq:kl-scaling} itself. First, it is the leading term of an asymptotic expansion: it requires $N_{\text{train}}$ large enough for the quadratic approximation \eqref{eq:kl-quadratic} to hold across the sampling distribution of $\hat\theta$, which fails when $P$ is not small compared to $N_{\text{train}}$ or when some directions become non-identifiable (e.g.\ degenerate mixture components); in that regime the effective number of parameters is smaller than $P$. Second, under misspecification the same expansion around the KL projection $\theta^{*}$ of $p_{0}$ onto the family yields a variance term $\mathrm{tr}(A^{-1}B)/(2N_{\text{train}})$, with $A$ the expected curvature and $B$ the score covariance ($A=B=F$ for a correct model), on top of the bias floor $\mathrm{KL}(p_{0}\,\|\,p_{\theta^{*}})$, which does not shrink with $N_{\text{train}}$. The floor eventually dominates the discrepancy, and the rejection point saturates at $N_{\text{test}}\sim 1/\mathrm{KL}(p_{0}\,\|\,p_{\theta^{*}})$ independently of further growth in $N_{\text{train}}$: this is the bias that uncertainties cannot absorb and that must be addressed at the level of model design.

These considerations sharpen the usage criteria of Fig.~\ref{fig:when}~\cite{Grosso:2026fgo}. When the fitted density enters an inference as the reference hypothesis, its estimation error may be neglected only while it lies below the sensitivity of that inference, and the validation test used to establish this must be at least as powerful as the downstream analysis (see the discussion of Sec.~\ref{sec:related}); otherwise, mismodeling that validation cannot see may still surface downstream as a false positive. Equation~\eqref{eq:ntest-scaling} bounds how long such safety can last for any analysis: beyond $N_\mathrm{test} \gtrsim N_\mathrm{train}/P$, even a perfectly specified model is rejected when tested, and the uncertainties of the fit must be incorporated as a composite null, as described in Sec.~\ref{subsec:gof}.

\section{Normalizing flows and the limits of ensemble uncertainties}
\label{app:nf}

Normalizing flows~\citep{NormalizingFlows} are among the most widely used and flexible neural density estimators in high-energy physics, both as surrogates for expensive simulation~\cite{MLreview,Krause:classifier-metric} and as background models~\cite{Hallin:2021wme,Raine:2022hht,Das:2024eie,Krause:2023uww,ATLAS:2026ugi}, including density estimation in learned embedding spaces for anomaly detection searches~\cite{Li:2026azw}.
Unlike the Gaussian mixtures of the main text, a flow has no tractable parameter covariance, so the score-based, full-parameter-space construction of Sec.~\ref{sec:method} is unavailable.

In this appendix, we consider the same two numerical experiments from Sec.~\ref{sec:exp}, a two-dimensional toy with a bimodal Gaussian and a skew-normal feature and a four-dimensional latent space for jet representations, and we ask the following two questions: (1) does the epistemic uncertainty of a normalizing flow ensemble admit a parametrization that a likelihood-ratio test can profile over? and (2) with this uncertainty propagated in the null, does a normalizing flow ensemble pass a composite goodness-of-fit test such as the one presented in the main text?

Ref.~\citep{badaracco2025an} provides such a parametrization through the maximum-likelihood $w_i f_i$ ensembling of Ref.~\citep{Sean}, in which independently trained flows are frozen and combined into a weighted mixture whose weights carry the uncertainty. This appendix extends that work in two directions: to more complex target distributions, and to a goodness-of-fit test built on the same construction as the main text.

We describe how we train the normalizing flow models in Sec.~\ref{app:nf-density}, how we estimate uncertainties with the $w_i f_i$ ensembling in Sec.~\ref{app:nf-wifi}, and how we incorporate the uncertainties in the goodness-of-fit test in Sec.~\ref{app:nf-gof}.
The results of our numerical experiments, presented in Sec.~\ref{app:nf-results}, show that a normalizing-flow ensemble admits such a weight parametrization, and that propagating it restores closure up to and beyond $N_{\mathrm{test}}\simeq N_{\mathrm{train}}$ for the largest ensemble ($M=128$) in the two-dimensional toy, but not in the four-dimensional embedding, where the propagated weight uncertainty is negligible compared to the flow's misspecification: the composite test collapses onto the point-null and the model is rejected well before $N_{\mathrm{test}}=N_{\mathrm{train}}$.

\subsection{Density estimation with normalizing flows}
\label{app:nf-density}
A normalizing flow learns an invertible, differentiable map $T_\phi$ from a simple base distribution to the target,
\begin{equation}
  f_\phi(x) \;=\; p_z\!\big(T_\phi^{-1}(x)\big)\,
  \left|\det \frac{\partial T_\phi^{-1}(x)}{\partial x}\right| ,
  \qquad p_z = \mathcal{N}(0,\mathbb{I}_D),
\end{equation}
and is trained with a maximum likelihood objective.
We use \texttt{nflows}~\citep{nflows} with masked rational-quadratic autoregressive layers, reverse permutations between layers, dropout $0.2$, and a standard normal base distribution. The architecture is chosen separately for each dataset by a grid search on held-out likelihood: four layers ($8$ bins, $4$ blocks, hidden width $64$) for the two-dimensional toy, and a larger flow with six layers ($10$ bins, $8$ blocks, hidden width $128$) for the four-dimensional jet embedding.
The training settings are the same for both datasets and were kept fixed in both grid searches: Adam~\citep{KingmaBa2015Adam} with batch size $512$ and an initial learning rate of $10^{-5}$, cosine-annealed to $10^{-6}$ over $1000$ epochs, with early stopping (patience $10$ epochs) on a $20\%$ validation split of each member's training sample. In both cases training ends by early stopping, typically after ${\sim}100$ epochs, well before the $1000$-epoch limit.

For each of the two problems (the two-dimensional toy and the four-dimensional jet embedding), we train a pool of $512$ flows, each an independent run on its own bootstrap replica of the $N_{\mathrm{train}}=10^{5}$ training set, so the spread across members reflects both data and optimization variability. An ensemble of size $M$ is built by drawing $M$ members uniformly at random from this pool, and we show $M=1,32,128$. 
Both the fitted density and its propagated weight uncertainty of Eq.~\eqref{eq:app-predvar} depend on which members happen to be drawn, so a single draw gives a noisy estimate of the uncertainty. To remove this dependence, we repeat the draw ten times, obtaining ten independent ensemble choices of size $M$, each fit independently (Sec.~\ref{app:nf-wifi}). In the marginal figures below the fitted model is the median over the ten choices and the band is the per-choice weight uncertainty averaged over them. This same per-choice uncertainty is what the goodness-of-fit test profiles over: it enters the null as the Gaussian prior $\pi(\boldsymbol u)=\mathcal N(\hat{\boldsymbol u},\Sigma_u)$ of Sec.~\ref{app:nf-gof}, so the band shown here is exactly the uncertainty propagated into the test. Each of the ten choices is then tested and calibrated independently, and the ten results are pooled into the reported $p$-value and its band (Sec.~\ref{app:nf-results}).

\subsection{Uncertainty quantification via \texorpdfstring{$w_i f_i$}{wifi} ensembling}
\label{app:nf-wifi}
Following~\citep{Sean}, we freeze the trained flows $f_i(x)$ as fixed basis functions and place the uncertainty on a small set of mixture weights $w_i$. The ensemble density is
\begin{equation}\label{eq:app-g}
  g(x) \;=\; \sum_{i=1}^{M} w_i\, f_i(x),
  \qquad w_M \;=\; 1 - \sum_{i=1}^{M-1} w_i ,
\end{equation}
where the sum-to-one constraint preserves normalization and leaves $M-1$ free weights $\boldsymbol u=(w_1,\dots,w_{M-1})$. The underlying assumption is that the true density lies in the span of $\{f_i\}$. The weights are signed ($w_i\ge0$ is not imposed because allowing cancellations between the networks improves expressivity) and are fit by minimizing the negative log-likelihood on the training sample,
\begin{equation}\label{eq:app-nll}
  \hat{\boldsymbol u}
  \;=\; \arg\min_{\boldsymbol u}\left[-\frac1N \sum_{n=1}^{N} \log g(x_n)\right],
\end{equation}
which is convex in $\boldsymbol u$ on the feasible set $\{\boldsymbol u: g(x_n)>0\ \forall n\}$. We find $\hat{\boldsymbol u}$ by damped Newton steps with a feasibility line search. The resulting estimator $\hat g(x)=g(x;\hat{\boldsymbol u})$ is the best estimate of the ensemble density.

Treating $\hat{\boldsymbol u}$ as an M-estimator~\citep{Huber}, its covariance is given by the sandwich formula~\citep{White1982}
\begin{equation}\label{eq:app-sandwich}
\begin{aligned}
  \mathrm{Cov}(\hat{\boldsymbol u}) &\;=\; \frac1N\, A^{-1} B\, A^{-1}, \\
  A &\;=\; \left.\frac{\partial^2 \mathcal L}{\partial \boldsymbol u\,\partial \boldsymbol u^\top}\right|_{\hat{\boldsymbol u}},
  \qquad
  B \;=\; \frac1N\sum_{n=1}^N s_n s_n^\top .
\end{aligned}
\end{equation}
Here $\mathcal L$ is the mean negative log-likelihood of Eq.~\eqref{eq:app-nll}, $A$ its Hessian at the optimum, and $B$ the covariance of the per-event scores $s_n$, whose components are the weight-difference sensitivities
$s_{n,j} = -\big(f_j(x_n)-f_M(x_n)\big)/g(x_n)$. We denote this $(M{-}1)\times(M{-}1)$ covariance by $\Sigma_u \equiv \mathrm{Cov}(\hat{\boldsymbol u})$. Propagating it to the density through the same sensitivities gives the pointwise predictive variance
\begin{equation}\label{eq:app-predvar}
  \sigma_{\hat g}^2(x)
  \;=\; \sum_{j,k=1}^{M-1}\big(f_j(x)-f_M(x)\big)\,
        \Sigma_{u,\,jk}\,
        \big(f_k(x)-f_M(x)\big),
\end{equation}
i.e.\ an uncertainty band $\hat g(x)\pm\sigma_{\hat g}(x)$ that supplies the pointwise epistemic uncertainty later propagated into the goodness-of-fit test of Sec.~\ref{app:nf-gof}.\footnote{In principle, as discussed in Ref.~\citep{Sean}, these sandwich uncertainties are conditional on the $f_i$ and only have proven coverage guarantees when the $w_i$ are fit on a distinct dataset from the $f_i$. In our experiments, we do not observe any significant effect from dataset reuse, so we present results from the simpler single dataset configuration.}

Equation~\eqref{eq:app-sandwich} is the exact analogue of the sandwich covariance discussed in the main text,\footnote{In the main text, we dismissed the sandwich estimator because it reduces to the standard inverse Hessian estimator where the model is well-specified. However, here we present the more general estimator because it is applicable to objectives using other ensemble parametrizations where the normalization of the model is not manifest, along the lines of Ref.~\cite{Sean}.} but restricted to the $M-1$ mixture weights: it captures only re-weightings $\sum_i \delta w_i\, f_i(x)$ of the fixed members, not deformations of any individual flow. As shown in Sec.~\ref{app:nf-results}, this is the reason why the composite test built on it approaches the uniform null $p$-value only in the 2D toy, and fails to improve on the point-null in the 4D embedding.

\subsection{Uncertainty-aware goodness-of-fit test}
\label{app:nf-gof}
We test the ensemble density with the composite-null likelihood-ratio construction of the main text (Sec.~\ref{subsec:gof}), now applied with the mixture weights as the profiled parameters. 

\paragraph{Null.}
The null is the family of re-weighted mixtures $g$ of Eq.~\eqref{eq:app-g}, with $\boldsymbol u$ constrained by the Gaussian prior centered on the fit,
\begin{equation}\label{eq:app-prior}
  \pi(\boldsymbol u) \;=\; \mathcal N\!\big(\boldsymbol u;\, \hat{\boldsymbol u},\,\Sigma_u\big).
\end{equation}
The trainable parameters $\boldsymbol u$ are profiled under this prior when the statistic is evaluated.

\paragraph{Alternative.}
The alternative enriches the null by a multiplicative exponential tilt along a fixed dictionary of localized kernels, as in the main text \citep{NonParametricMethods},
\begin{align}
  h_{\boldsymbol u,\boldsymbol\alpha}(x)
  &\;=\; \frac{g_{\boldsymbol u}(x)\,
          \exp\!\big(\textstyle\sum_{j=1}^{J}\alpha_j\, k_j(x)\big)}
         {Z(\boldsymbol u,\boldsymbol\alpha)},
  \label{eq:app-alt}\\[2pt]
  Z(\boldsymbol u,\boldsymbol\alpha)
  &\;=\; \int g_{\boldsymbol u}(x)\,
         \exp\!\big(\textstyle\sum_{j=1}^J \alpha_j k_j(x)\big)\, dx ,
  \label{eq:app-Z}
\end{align}
with Gaussian kernels $k_j(x)=(2\pi\sigma^2)^{-D/2}\exp\!\big(-\|x-c_j\|^2/2\sigma^2\big)$. 
The kernel dictionary is similar to that of the main text (Sec.~\ref{subsec:gof}), using $J$ Gaussian kernels of a single width $\sigma$ (values in Sec.~\ref{app:nf-results}), but in this case centered on the first $J$ points of the sample under test. Since the $J=500$ kernels are each about as wide as the spread of the data and overlap heavily, so every region populated by the data is covered whichever way the centers are placed, and we expect the choice of placement to have no significant effect on the test.
Setting $\boldsymbol\alpha=0$ recovers the null ($h=g$), so the null is nested in the alternative. The exponential tilt preserves the sign of $g_{\boldsymbol u}$, and $h$ is normalized up to the Monte Carlo estimate $\widehat Z$ of Eq.~\eqref{eq:app-Zhat}.

As with the test in the main text, two regularizations act on the tilt: an $L_2$ ridge $\tfrac12\lambda\|\boldsymbol\alpha\|^2$ and a box $|\alpha_j|\le A$ on the tilt coefficients. The normalization~\eqref{eq:app-Z} is estimated by importance sampling from the equal-weight member mixture $q(x)=\tfrac1M\sum_m f_m(x)$,
\begin{equation}\label{eq:app-Zhat}
  \widehat Z(\boldsymbol u,\boldsymbol\alpha)
  = \frac1R\sum_{r=1}^{R}\frac{g_{\boldsymbol u}(y_r)}{q(y_r)}\,
        \exp\!\big(\textstyle\sum_j \alpha_j k_j(y_r)\big),
\end{equation}
with $y_r \sim q$ and $R=N_{\mathrm{ref}}=N_{\mathrm{test}}$ reference points. The reference weights are self-normalized so that $\widehat Z(\boldsymbol u,\boldsymbol 0)=1$.

\paragraph{Test statistic.}
The test statistic follows the construction of the main text (Sec.~\ref{subsec:gof}), with the mixture weights $\boldsymbol u$ in place of the whitened coordinates $v$ and the prior $\pi(\boldsymbol u)$ in place of the Gaussian constraint on $v$. Following the standard LHC nuisance-parameter prescription~\citep{Cowan:2010js}, we define the penalized log-likelihood
\begin{equation}\label{eq:app-ell}
  \ell(\boldsymbol u,\boldsymbol\alpha) = \log\mathcal L(\mathcal D\mid h_{\boldsymbol u,\boldsymbol\alpha}) + \log\pi(\boldsymbol u) - \tfrac{\lambda}{2}\|\boldsymbol\alpha\|^2 .
\end{equation}
The composite statistic is the profile likelihood ratio
\begin{equation}\label{eq:app-lrt}
  t_{\rm comp}(\mathcal D) = 2\Big[\max_{\boldsymbol u,\boldsymbol\alpha} \ell(\boldsymbol u,\boldsymbol\alpha) - \max_{\boldsymbol u} \ell(\boldsymbol u,0)\Big] ,
\end{equation}
where the second term lets the ensemble move within its weight uncertainty to best fit the data, and the first additionally switches on the localized kernels, so that only the discrepancy the weights cannot absorb contributes. As in the main text, the point-null statistic fixes the weights at their best fit,
\begin{equation}\label{eq:app-tpoint}
  t_{\rm point}(\mathcal D) = 2\Big[\max_{\boldsymbol\alpha} \ell(\hat{\boldsymbol u},\boldsymbol\alpha) - \ell(\hat{\boldsymbol u},0)\Big] .
\end{equation}

\paragraph{Calibration.}
The training data (used to fit the flows and the weights $\hat{\boldsymbol u}$) are kept strictly separate from the sample $\mathcal D$ on which $t$ is evaluated. The null distribution of $t$ is built from $B$ pseudo-experiments. We draw posterior-predictive toys: each toy draws a weight vector $\boldsymbol u^{(b)}\sim\pi$ and then a dataset $\mathcal D^{(b)}\sim g_{\boldsymbol u^{(b)}}$, so that the null itself carries the weight uncertainty. 
Because the weight fit of Eq.~\eqref{eq:app-nll} is convex, unlike the non-convex mixture fit of the main text, whose toys must be re-fit from a cold start, re-fitting each toy would recover a $\boldsymbol u^{(b)}$ in the vicinity of $\hat{\boldsymbol u}$ regardless of initialization. We then expect drawing $\boldsymbol u^{(b)}\sim\pi$ to perform similarly, so we do not re-fit the data-generating weights per toy and thus only show the Gaussian calibration. A full bootstrap calibration which refits the whole ensemble of flows in every pseudo-experiment could outperform this procedure, but would be computationally prohibitive.

Because the weights are signed, $g_{\boldsymbol u}$ is not a probability mixture and cannot be sampled member-by-member. We instead draw toys by sampling-importance-resampling (SIR) from the positive proposal $q$. The statistic is then evaluated on each toy exactly as on the data, re-maximizing over $\boldsymbol u$ and $\boldsymbol\alpha$ as in Eq.~\eqref{eq:app-lrt}, giving $\{t^{(b)}\}$, and the $p$-value of the observed statistic is the empirical tail probability $p=(\#\{t^{(b)}\ge t(\mathcal D)\}+1)/(B+1)$. 
For the point-null statistic, toys are drawn from $g_{\hat{\boldsymbol u}}$ without sampling the weights. Because the tilt is regularized (ridge and box) and its kernel dictionary is strongly collinear, neither statistic follows the asymptotic Wilks $\chi^2$, whose number of degrees of freedom would equal the number of tilt coefficients; the effective number of degrees of freedom is much smaller. We therefore do not rely on Wilks' theorem, but calibrate the statistics empirically from the pseudo-experiments, against which all $p$-values in Sec.~\ref{app:nf-results} are evaluated.

\subsection{Results}
\label{app:nf-results}
We apply the construction of Secs.~\ref{app:nf-density}--\ref{app:nf-gof} to the two-dimensional toy of Sec.~\ref{subsec:toy} ($N_{\mathrm{train}}=10^5$), for ensembles of $M$ flows up to $M=128$. The tests use the kernel dictionary of the main text ($J=500$,
$\sigma=0.4$), ridge $\lambda=1$ and box $A=3$. Each null is calibrated on $B=100$ pseudo-experiments and the $p$-value is estimated from $100$ draws of $N_{\mathrm{test}}$ events from the truth.
We then repeat the construction on the four-dimensional jet embedding of Sec.~\ref{subsec:jets}, with the same $N_{\mathrm{train}}$, ensemble sizes and test settings, and the four-dimensional kernel width of the main text, $\sigma=0.53$.

\begin{figure*}[htbp]
  \centering
  \includegraphics[width=\linewidth]{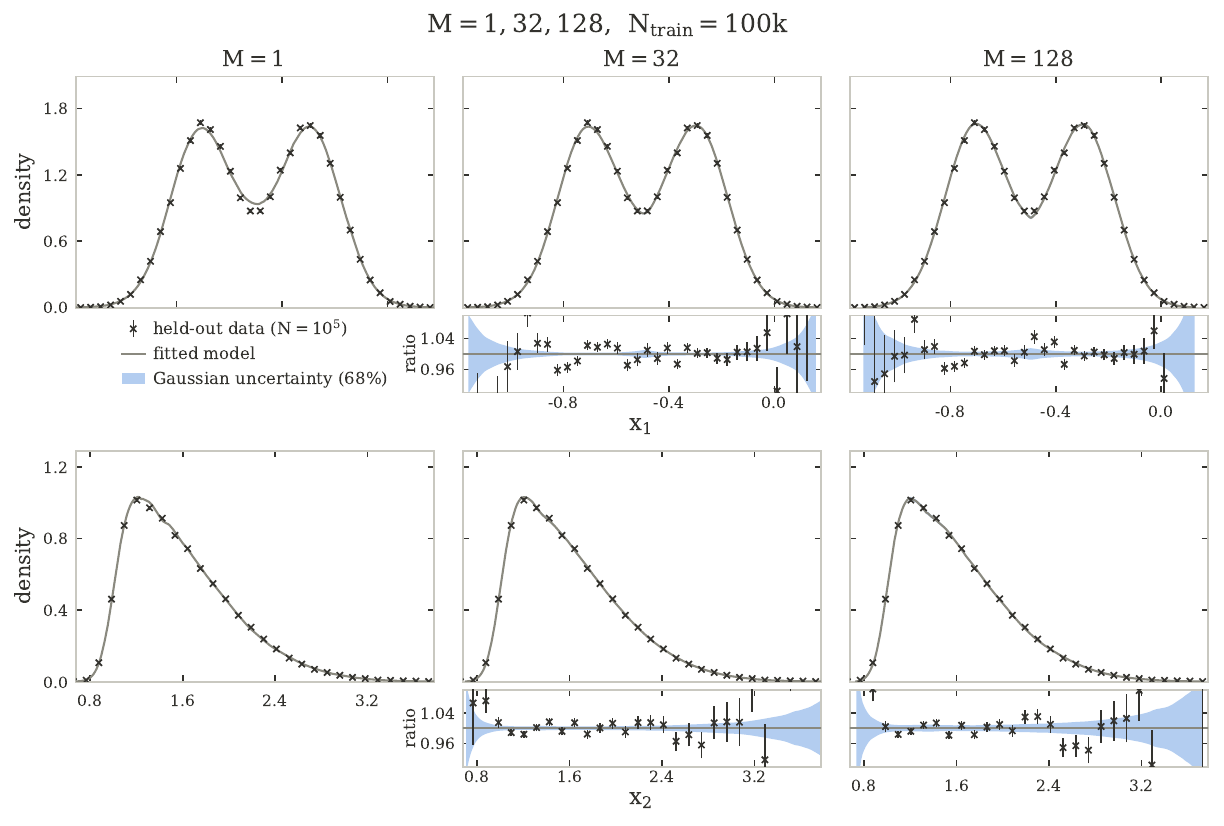}
  \caption{Two-dimensional toy: fitted marginals of features $x_1$ (top) and $x_2$
  (bottom) for ensembles of $M=1,32,128$ flows (columns). Points are held-out data
  ($N=10^5$); the gray line is the fitted ensemble density
  $\hat g=\sum_i \hat w_i f_i$, taken as the median over the ten ensemble choices
  (Sec.~\ref{app:nf-density}); the blue band is the $68\%$ weight-uncertainty
  $\hat g\pm\sigma_{\hat g}$ of Eq.~\eqref{eq:app-predvar}, averaged over the ten
  choices and shown in the ratio-to-fit sub-panels. The single flow ($M=1$) carries
  no fitted weights and hence no band. As $M$ grows the fit tightens onto the target
  (most visibly the bimodal valley of $x_1$) and the band, narrow in the bulk, widens
  in the low-statistics tails.}
  \label{fig:nf-marg-2d}
\end{figure*}

\begin{figure*}[htbp]
  \centering
  \includegraphics[width=\linewidth]{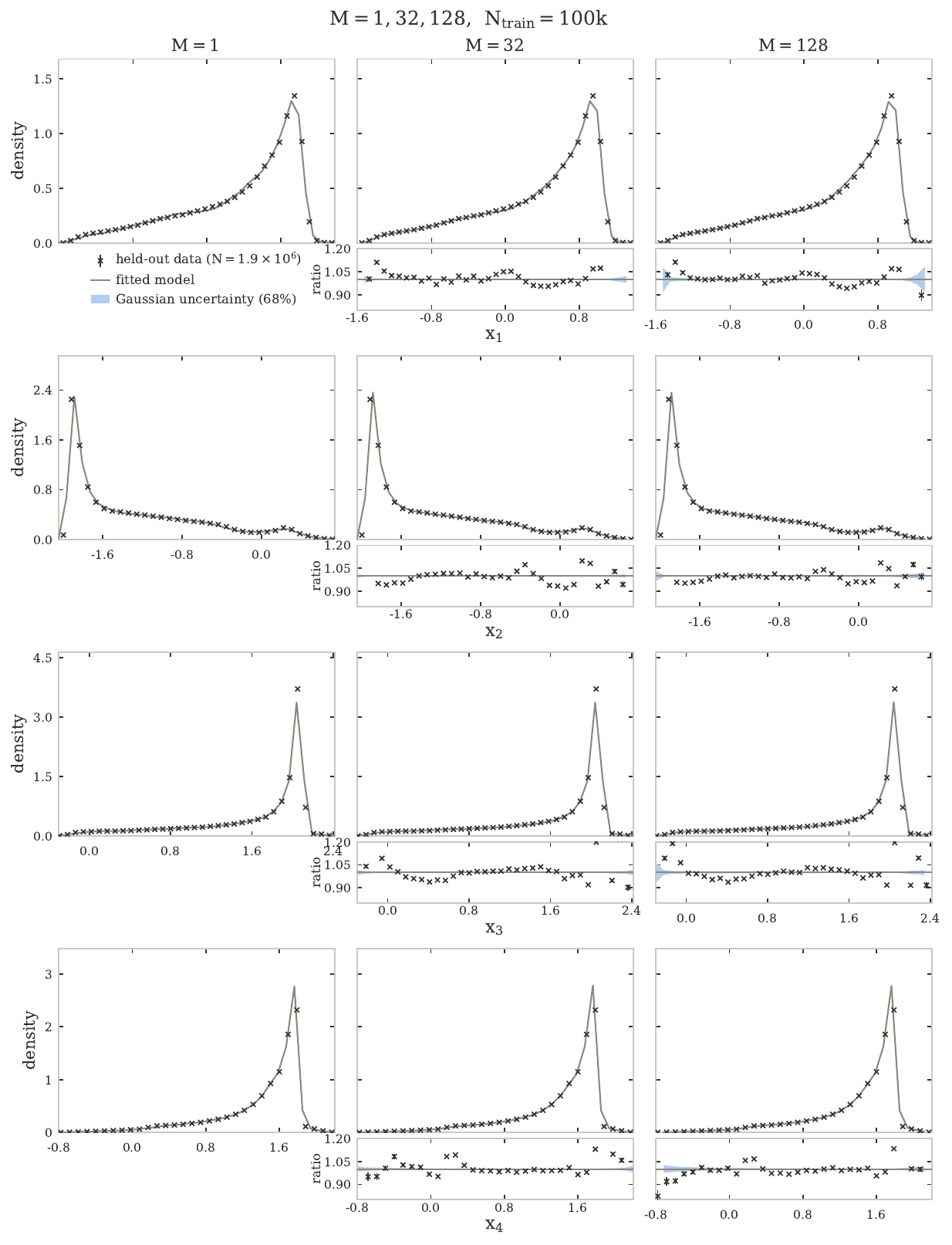}
  \caption{Four-dimensional jet embedding: fitted marginals of $x_1$--$x_4$ (rows) for $M=1,32,128$ flows (columns), constructed as in Fig.~\ref{fig:nf-marg-2d}; held-out data $N=1.9\times10^6$. The fitted marginals deviate from the data around the sharp peaks, while the weight-uncertainty band remains negligible at every $M$, so the propagated uncertainty does not cover the discrepancy.}
  \label{fig:nf-marg-4d}
\end{figure*}

\paragraph{Density quality.}
Figure~\ref{fig:nf-marg-2d} shows the fitted two-dimensional marginals for $M=1,32,128$. A single flow ($M=1$), which carries no parametrized uncertainty, reproduces the bulk of the target but visibly mismodels the peaks and the valley of the bimodal feature~1. As $M$ increases the ensemble median tightens onto the target (most visibly the bimodal valley of feature~1) and a weight-uncertainty band develops, narrow in the bulk and widening in the low-statistics tails. In the four-dimensional embedding (Fig.~\ref{fig:nf-marg-4d}) the picture is different: the fitted marginals show structured deviations from the held-out data, most visibly around the sharp peaks, while the weight-uncertainty band remains negligibly small at every $M$. The propagated uncertainty therefore does not cover the discrepancy between the ensemble and the data, which anticipates the goodness-of-fit results below.

\begin{figure*}[htbp]
  \centering
  \includegraphics[width=0.7\linewidth]{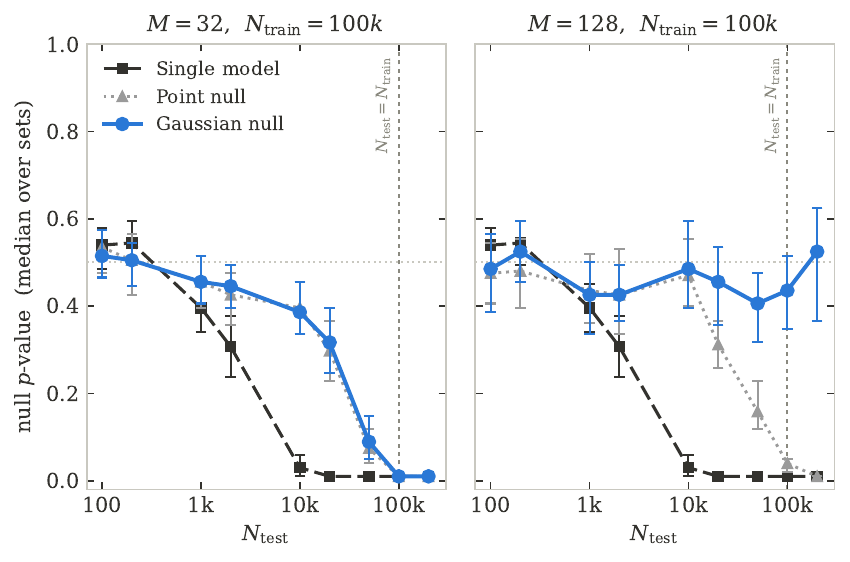}
  \caption{\textbf{Closure test for the 2D toy problem with NF ensemble.} Median test $p$-value versus $N_{\mathrm{test}}$ for
  ensembles of $M=32$ (left) and $M=128$ (right) flows, $N_{\mathrm{train}}=10^5$. Each panel shows the single-flow baseline (black), the point-null (gray), and the composite
  ``Gaussian'' null that propagates the weight uncertainty (blue). Error bars are the
  $95\%$ confidence interval of the pooled median, from a hierarchical bootstrap over the
  ten ensemble choices and their null/test toys; the dotted line marks the nominal
  $p=0.5$ and the dashed vertical $N_{\mathrm{test}}=N_{\mathrm{train}}$. Only at $M=128$
  does the composite test stay near nominal up to and beyond $N_{\mathrm{test}}=N_{\mathrm{train}}$.}
  \label{fig:nf-pv-2d}
\end{figure*}

\paragraph{Goodness-of-fit.}
Fig.~\ref{fig:nf-pv-2d} shows the median test $p$-value versus $N_{\mathrm{test}}$ for $M=32,128$. As expected, the test sensitivity increases with $N_{\mathrm{test}}$: the $p$-value stays near nominal while the test sample is too small to resolve the finite-$N_{\mathrm{train}}$ error of the fit, and falls once it does. The single flow degrades earliest, by $N_{\mathrm{test}}\sim10^4$, as anticipated in Sec.~\ref{sec:motivation}: a flow's large parameter count and imperfect inductive bias push the rejection onset to $N_{\mathrm{test}}\ll N_{\mathrm{train}}$. A single flow is therefore not a valid reference model at these test sizes. Ensembling and propagating the weight uncertainty push this onset outward, and the benefit grows with $M$. A moderate ensemble ($M=32$) delays the degradation to $N_{\mathrm{test}}\sim2$--$5\times10^4$ but still reaches the floor $1/(B+1)$ by $N_{\mathrm{test}}=N_{\mathrm{train}}=10^5$ with little difference between the composite and point-null (median $p\approx0.09$ vs $0.07$ at $N_{\mathrm{test}}=5\times10^4$). This happens although the $M=32$ ensemble closely reproduces the two-dimensional marginals (Fig.~\ref{fig:nf-marg-2d}): the test probes the joint density, where the residual error of the fit is resolved. Only at $M=128$ does propagating the weight uncertainty make a clear difference: the composite test stays near nominal, $p\approx0.43$, whereas the point-null falls to $p\approx0.16$ at $N_{\mathrm{test}}=5\times10^4$ and to the floor by $N_{\mathrm{test}}=N_{\mathrm{train}}$. With $M=128$ the composite test gives $p\approx0.43$ out to $N_{\mathrm{test}}=2\times10^5$, i.e.\ it does not reject the model even in the genuine oversampling regime $N_{\mathrm{test}}>N_{\mathrm{train}}$.

\begin{figure*}[htbp]
  \centering
  \includegraphics[width=0.7\linewidth]{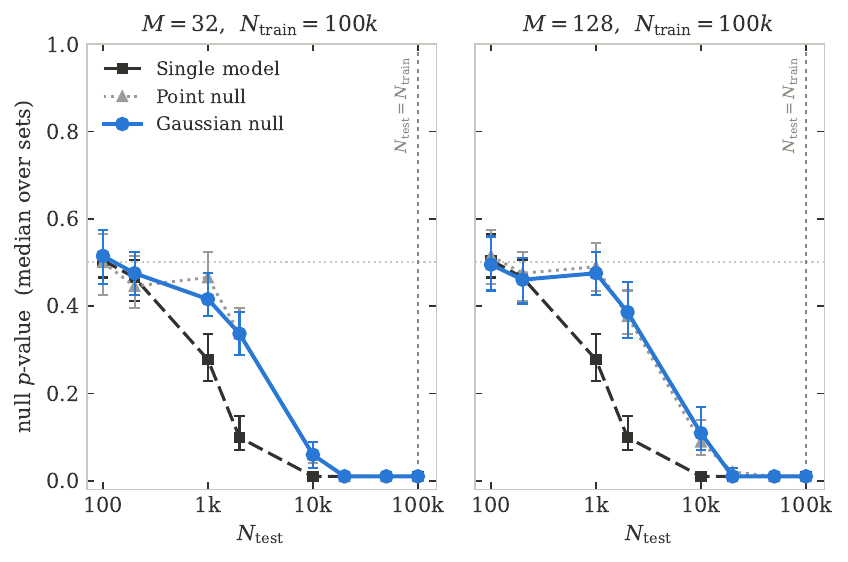}
  \caption{\textbf{Closure test for the 4D particle jets problem with NF ensemble.}  Median test $p$-value versus $N_{\mathrm{test}}$ for $M=32$ (left) and $M=128$ (right) flows, constructed as in Fig.~\ref{fig:nf-pv-2d}. At every $M$ the composite (Gaussian) null tracks the point-null to within the pseudo-experiment scatter: propagating the weight uncertainty does not help, and all constructions reach the floor $1/(B+1)$ by $N_{\mathrm{test}}=N_{\mathrm{train}}=10^5$.}
  \label{fig:nf-pv-4d}
\end{figure*}

In four dimensions the picture changes (Fig.~\ref{fig:nf-pv-4d}). The single flow degrades by $N_{\mathrm{test}}\sim2\times10^3$ and the ensembles by $N_{\mathrm{test}}\sim10^4$. There is no improvement from propagating the weight uncertainty: the composite $p$-value is inseparable from the point-null within the pseudo-experiment variations at every $M$ and every $N_{\mathrm{test}}$, with no separation even at $M=128$. All constructions reach the floor $1/(B+1)$ by $N_{\mathrm{test}}\sim2\times10^4$, well below $N_{\mathrm{train}}=10^5$, and even $M=128$ is fully rejected.

\paragraph{Discussion.}
Since every null is calibrated with pseudo-experiments drawn from the composite-null itself, the loss of $p$-value at large $N_{\mathrm{test}}$ reflects a failure of the ensemble uncertainty to cover the error of the fit, not an artifact of the statistic.
Ensembling the flows and propagating the weight uncertainty push this rejection outward in the 2D toy study, and with $M=128$ the model survives up to $N_{\mathrm{test}}\simeq N_{\mathrm{train}}$ and into the oversampling regime $N_{\mathrm{test}}>N_{\mathrm{train}}$ probed here, up to $N_{\mathrm{test}}=2N_{\mathrm{train}}$. In the more complex 4D problem, the construction fails to close: ensembling still improves, but the weight covariance $\Sigma_u$ spans only re-weightings $\sum_i \delta w_i f_i(x)$ of the frozen members, and in the embedding the resulting uncertainty is negligible compared to the discrepancy between the ensemble and the data. With an uncertainty this small, the profiled weights can barely move, the composite null effectively reduces to the point-null, and the two statistics coincide: even an ensemble of $M=128$ flows is rejected well before $N_{\mathrm{test}}=N_{\mathrm{train}}$. This is the limitation anticipated in Sec.~\ref{app:nf-wifi}: re-weighting frozen flows can make up for the error of a simple target, but not for that of a more complex one.

\section{Implementation details}
\label{app:implementation}

The nominal fit and every bootstrap refit use EM with a $k$-means initialization (specifically, the \verb|k-means++| initialization implemented in the \verb|GaussianMixture| class in scikit-learn~\cite{scikit-learn}). The scores of Eq.~\eqref{eq:score} can be computed analytically or by central finite differences,
\begin{equation}
  \psi_a(x) \approx \frac{\log p_{\hat\theta+\epsilon e_a}(x) - \log p_{\hat\theta-\epsilon e_a}(x)}{2\epsilon},
\end{equation}
which costs $2P$ evaluations of the log-density on each sample.
The normalization of the tilted model for the alternative hypothesis, as well as the standardization of the features, are calculated using reference samples drawn from a pool of size $2 \times 10^6$ from the best-fit \methodname model.
$10 N_\text{test}$ samples are used to compute the normalization to ensure that its uncertainty is subdominant, and $10^6 = 10 N_\text{train}$ samples are used to compute the Fisher matrix.
The feature standardization is performed with the whole pool.

The $J=500$ kernel centers $c_j$ are obtained by $k$-means clustering of a subsample of $R_0$. The width $\sigma$ of the kernels is taken to be uniform and isotropic, with $\sigma = 0.4$ in the two-dimensional case study and $\sigma = 0.53$ in the four-dimensional case study.

The kernel amplitudes carry the ridge penalty $\lambda_\alpha\|\alpha\|^2$ of Eq.~\eqref{eq:ell} and are restricted to
\begin{equation}
  |\alpha_j| \le b_j = \frac{A}{\max_{R_0} |\tilde k_j|},
  \label{eq:clip}
\end{equation}
which bounds the log-density distortion of each kernel by $A$. We use $A=10$, i.e., local distortions within a factor $e^{\pm 10}$. Both regularizations are applied identically to the data and to every pseudo-experiment. $\lambda_\alpha$ is taken to be $1$ throughout.

\begin{figure*}[t]
    \centering
    \includegraphics[width=\linewidth]{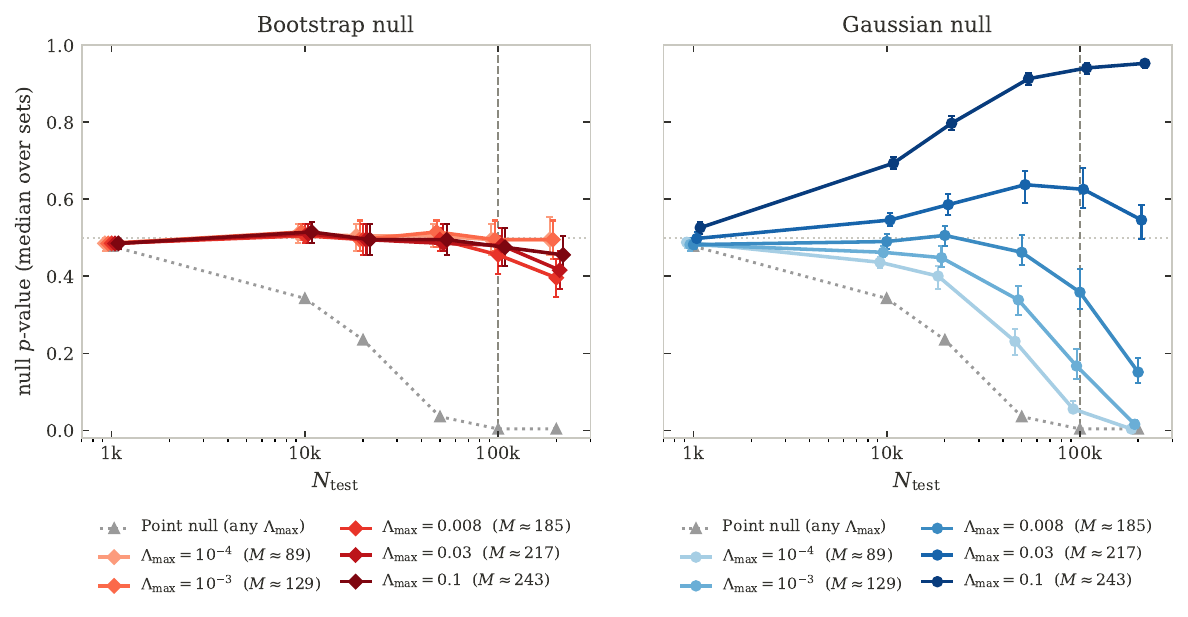}
    \caption{\textbf{$\Lambda_\text{max}$ scan for the 2D case study.}  Median test $p$-value versus $N_{\mathrm{test}}$
  for $K=64$ \methodname with the bootstrap (left) and Gaussian (right) calibration prescription. The bootstrap calibration is largely robust to $\Lambda_\text{max}$, while the results for the Gaussian calibration vary strongly with $\Lambda_\text{max}$. Each shade corresponds to a different value of $\Lambda_\text{max}$, and the point null is independent of this hyperparameter.}
    \label{fig:lam_ablation}
\end{figure*}

The $\Lambda_\text{max}$ hyperparameter is taken to be $0.008$ in all of the studies in the main body, and Fig.~\ref{fig:lam_ablation} shows the effects of changing $\Lambda_\text{max}$ in the 2D example with $K=64$.
As anticipated in the main body, the performance of the Gaussian-calibrated null is strongly dependent on the choice of $\Lambda_\text{max}$, since its sampling from the null hypothesis is directly dependent on $\Lambda_\text{max}$, while the bootstrap-calibrated null is more robust, as $\Lambda_\text{max}$ only affects the definition of the test statistic and not the sampling of the null.

\section{Null distribution of the test statistic}
\label{app:null-distribution}
We show that, to second order in $(v,\alpha)$ and for $\lambda_\alpha = 0$, $t_{\rm comp}$ follows a $\chi^2_J$ distribution under the Gaussian composite null. Let $f = (\Phi, \tilde k) \in \mathbb{R}^{M+J}$ denote the full feature vector, which has zero mean under $p_{\hat\theta}$ and, by Eq.~\eqref{eq:unit-variance}, covariance
\begin{equation}
  \Sigma = \begin{pmatrix} \mathbb{I}_M / N_{\rm train} & C_{v\alpha} \\ C_{\alpha v} & C_{\alpha\alpha} \end{pmatrix},
\end{equation}
with $C_{\alpha\alpha} = \mathrm{Cov}[\tilde k]$ and $C_{\alpha v} = C_{v\alpha}^\top = \mathrm{Cov}[\tilde k, \Phi]$. Write $N \equiv N_{\rm test}$, $a \equiv N_{\rm test}/N_{\rm train}$, $u = (v,\alpha)$, and $s = \sum_i f(x_i) = (s_v, s_\alpha)$ for the feature sums over the test sample. Expanding $\log\widehat Z(u) = \tfrac12 u^\top \Sigma u + O(\|u\|^3)$ gives
\begin{equation}
  \ell(u) \simeq u^\top s - \tfrac12 u^\top H u,\,
  H = \begin{pmatrix} (1+a)\,\mathbb{I}_M & N C_{v\alpha} \\ N C_{\alpha v} & N C_{\alpha\alpha} \end{pmatrix},
\end{equation}
where the identity in the upper block comes from the Gaussian constraint. The numerator and denominator maximizations give $s^\top H^{-1} s / 2$ and $\|s_v\|^2 / 2(1+a)$, respectively, and the block-inverse identity yields
\begin{equation}
  t_{\rm comp} = s^\top H^{-1} s - \frac{\|s_v\|^2}{1+a} = r^\top S^{-1} r,
\end{equation}
with
\begin{align}
  r &= s_\alpha - \frac{N}{1+a}\, C_{\alpha v}\, s_v, \\
  S &= N C_{\alpha\alpha} - \frac{N^2}{1+a}\, C_{\alpha v} C_{v\alpha}.
\end{align}
Here $r$ is the vector of kernel sums with their projection on the score sums removed, and $S$ is the Schur complement of the $v$-block of $H$.

Under the Gaussian composite null, the test sample is drawn from $p_{\hat\theta+Wz} \simeq g_z$ with $z\sim\mathcal{N}(0,\mathbb{I}_M)$. The tilt shifts the mean of $f$ by $\Sigma_{\cdot v}\, z = (z/N_{\rm train},\, C_{\alpha v} z)$, so that, including the spread of $z$,
\begin{align}
  \mathrm{Cov}(s_v) &= a(1+a)\,\mathbb{I}_M, \\
  \mathrm{Cov}(s_\alpha, s_v) &= N(1+a)\, C_{\alpha v}, \\
  \mathrm{Cov}(s_\alpha) &= N C_{\alpha\alpha} + N^2 C_{\alpha v} C_{v\alpha}.
\end{align}
Substituting gives $\mathrm{Cov}(r) = S$. Since $r$ is asymptotically Gaussian, $t_{\rm comp} \sim \chi^2_J$, or $\chi^2$ with $\mathrm{rank}(S)$ degrees of freedom if the kernels are linearly dependent on $R_0$. The result is independent of $a$ and of the correlations $C_{\alpha v}$. The $M$ standardized directions drop out because the constraint on $v$ matches the distribution of $z$.

A ridge $\lambda_\alpha > 0$ adds $2\lambda_\alpha \mathbb{I}_J$ to the $\alpha$-block of $H$, which leaves $r$ unchanged and replaces $S$ by $S + 2\lambda_\alpha\mathbb{I}_J$ in the quadratic form. The statistic is then a weighted sum of $J$ independent $\chi^2_1$ variables with weights $\sigma_j/(\sigma_j + 2\lambda_\alpha) < 1$, where $\sigma_j$ are the eigenvalues of $S$. For $t_{\rm point}$, with $v \equiv 0$ and data drawn from $p_{\hat\theta}$, the same argument gives $t_{\rm point} = s_\alpha^\top (N C_{\alpha\alpha})^{-1} s_\alpha \sim \chi^2_J$.

\begin{figure*}
    \centering
    \includegraphics[width=\textwidth]{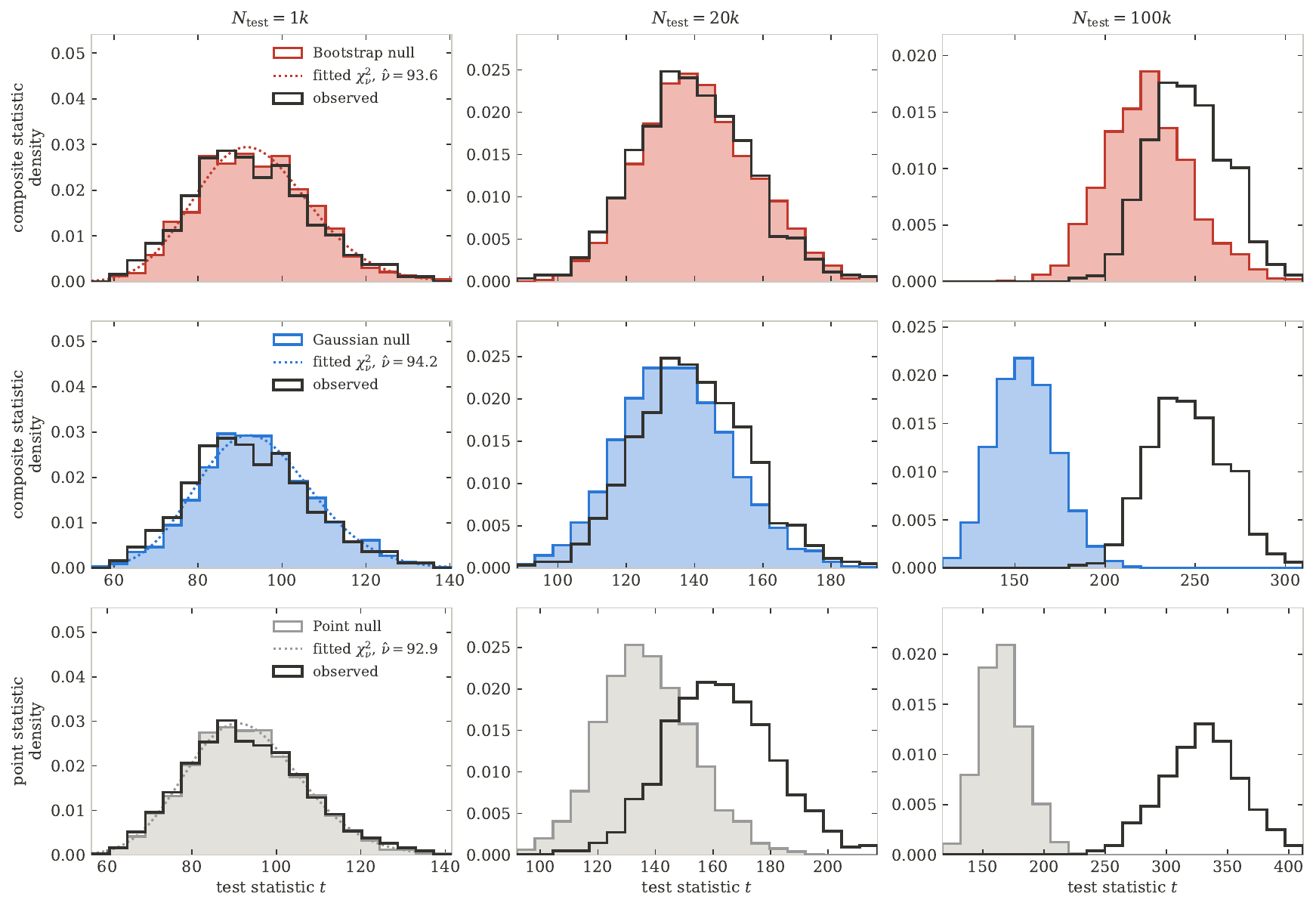}
    \caption{\textbf{Empirical distributions of the test statistics for the 4D particle jets problem}, for a GMM with $K=256$ fitted to $N_{\rm train}=10^{5}$ points. Columns: test-set size $N_{\rm test}=1\,000$, $20\,000$, $100\,000$. Rows: composite statistic with bootstrap calibration (red), composite statistic with Gaussian calibration (blue), point-null statistic (gray). Filled histograms: empirical null from $2\,500$ pseudo-experiments for Gaussian and point-null and $1\,000$ pseudo-experiments for bootstrap; black outline: observed statistic on 1 000 test sets, 100 drawn from the held-out pool for each of 10 independently trained models; gray dotted: best-fit $\chi^2_{\hat\nu}$.}
    \label{fig:null-distributions-4d}
\end{figure*}

Figure~\ref{fig:null-distributions-4d} examines the empirical null distributions for the four-dimensional particle-jets problem of Sec.~\ref{subsec:jets} with $K=256$ and $N_{\rm train}=10^{5}$, for the three null prescriptions of Sec.~\ref{subsec:tangent-calibration} and increasing test-set size. 
At $N_\text{test} = 1$k, all three empirical nulls agree with each other and with the observed statistic, and a $\chi^2_{\hat\nu}$ fitted to each null returns $\hat\nu \approx 93$, consistent with the predicted effective $\nu \approx 82$ and the presence of additional noise due to the finite sample used to estimate $Z$.
In the small-sample regime, the asymptotic approximation holds and the test cannot resolve the finite-sample uncertainty, and the three calibrations are interchangeable.

The tests separate as $N_\text{test}$ grows.
At $N_\text{test}=20$k, the two composite nulls still nearly reproduce the observed distribution (with bootstrap somewhat outperforming Gaussian), while the point-null distribution has already shifted below that of the observed test statistic.
At $N_\text{test} = 10^5 = N_\text{train}$, all three null distributions fall below the observed test statistic distribution (consistent with the $p$-values in Fig.~\ref{fig:closure-4d-100k}), but the point-null and Gaussian prescription fail dramatically while the bootstrap calibration remains fairly close to the observed distribution.

\section{Effects of kernel width}
\label{app:resolution}
\begin{figure*}[t]
    \centering
    \includegraphics[width=0.75\linewidth]{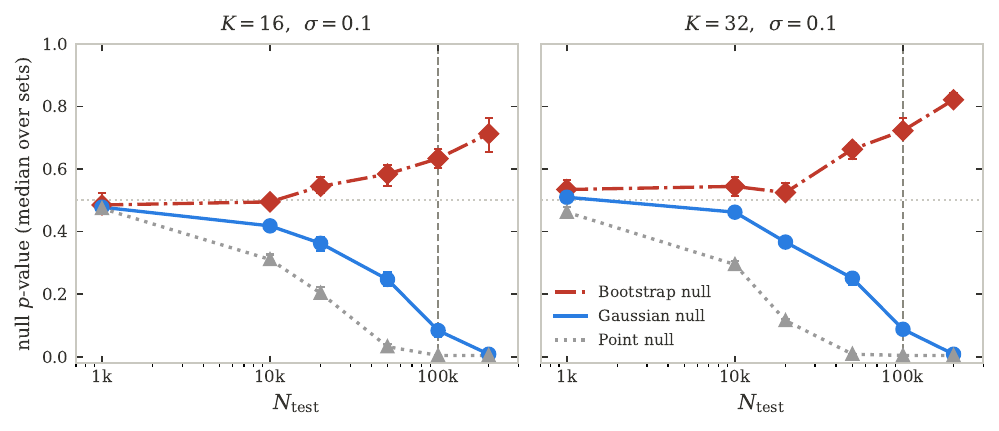}
    \caption{\textbf{Closure test for the 2D toy problem with $\sigma=0.1$.} As with Fig.~\ref{fig:closure-2d-100k}, median null $p$-value for the three goodness-of-fit tests of Sec.~\ref{subsec:gof} versus test-set size $N_\text{test}$ for $K=16,32$. Unlike Fig.~\ref{fig:closure-2d-100k}, however, the width of the test kernels is taken to be $\sigma=0.1$, which is substantially finer than the natural scale of the data.}
    \label{fig:small_sigma_closure}
\end{figure*}

In this appendix, we study the effects of the choice of kernel width $\sigma$ on the behavior of the goodness-of-fit test.
As discussed in the main body, we find that choosing $\sigma$ too small relative to the length scale of the data results in spurious calibration failure, establishing a tradeoff where $\sigma$ must be chosen large enough for the test to be reliable, and small enough for it to be sensitive.
In Fig.~\ref{fig:small_sigma_closure}, we show results of the closure test for the two-dimensional example for $K=16,32$ with all parameters chosen as in the main body, except $\sigma=0.1$ instead of $0.4$.
As with the larger kernels, the point null is insufficient and is rejected for all but the smallest values of $N_\text{test}$; the Gaussian null outperforms the point null, but is also miscalibrated for $N_\text{test} \geq 10$k.
On the other hand, the bootstrap-calibrated null deviates from closure in the opposite way: the median $p$-value is greater than $0.5$ at large $N_\text{test}$, indicating that the distribution of the test statistic is shifted rightward in calibration pseudo-experiments relative to the distribution in observed data.
In other words, the test finds more discrepancy, on average, with the best-fit trained model in pseudo-experiments (where that model is guaranteed to be well-specified) than in the observed data (where it is not), so this deviation from calibration is clearly spurious.
We ascribe this to an interplay between the structure of the \methodname model, which is a GMM, and the test.
When trained on a finite training set without strong regularization, GMMs generally overfit by placing narrow components in regions of feature space with disproportionately many events.
This overfitting is not pathological in our context, as it is part of the variance we aim to capture with uncertainties (as opposed to model bias, which we cannot capture; strong regularization would induce such bias).
The fine-grained structure induced by overfitting is then generically present in our trained models but not in the data-generating process, so an alternative hypothesis comprised of small-length-scale Gaussian functions will systematically fit a dataset drawn from a trained \methodname model better than it would observed data.
As such, we do not recommend using \methodname models in conjunction with a goodness-of-fit test sensitive to length scales substantially smaller than the natural length scales of the data, hence our choice of $\sigma$ in the main body to be comparable to the data standard deviation.

\clearpage

\bibliography{bibliography}

@inproceedings{badaracco2025an,
    title={An end-to-end pipeline for uncertainty-aware validation of generative {AI}},
    author={Giada Badaracco and Sean Benevedes and Gaia Grosso and Christina Reissel and {\AA}rrestad, Thea K. and Philip Harris},
    booktitle={Machine Learning and the Physical Sciences Workshop @ NeurIPS 2025},
    year={2025},
    url={https://openreview.net/forum?id=4J31chAHT7}
}

@article{Metzger:2025ecl,
    author = "Metzger, Kyle and Xu, Lana and Sodini, Mia and {\AA}rrestad, Thea K. and Govorkova, Katya and Grosso, Gaia and Harris, Philip",
    title = "{Anomaly-preserving contrastive neural embeddings for end-to-end model-independent searches at the LHC}",
    eprint = "2502.15926",
    archivePrefix = "arXiv",
    primaryClass = "hep-ex",
    doi = "10.1103/5n77-ynsp",
    journal = "Phys. Rev. D",
    volume = "112",
    number = "7",
    pages = "072011",
    year = "2025"
}

@inproceedings{
    bright2025autoscidact,
    title={{AutoSciDACT: Automated scientific discovery through contrastive embedding and hypothesis testing}},
    author={Samuel Bright-Thonney and Christina Reissel and Gaia Grosso and Nathaniel S. Woodward and Katya Govorkova and Andrzej Novak and Sang Eon Park and Eric A. Moreno and Philip Harris},
    booktitle={The Thirty-ninth Annual Conference on Neural Information Processing Systems},
    year={2025},
    eprint        = "2510.21935",
    archivePrefix = "arXiv",
    primaryClass  = "cs.LG",
    url={https://openreview.net/forum?id=vKyiv67VWa}
}

@article{Li:2026azw,
    author = "Li, Runze and Nachman, Benjamin and Noll, Dennis",
    title = "{Signal-Aware Contrastive Latent Spaces for Anomaly Detection}",
    eprint = "2603.25794",
    archivePrefix = "arXiv",
    journal = "",
    primaryClass = "hep-ph",
    month = "3",
    year = "2026"
}

@article{scikit-learn,
  title={Scikit-learn: Machine learning in {P}ython},
  author={Pedregosa, F. and Varoquaux, G. and Gramfort, A. and Michel, V.
          and Thirion, B. and Grisel, O. and Blondel, M. and Prettenhofer, P.
          and Weiss, R. and Dubourg, V. and Vanderplas, J. and Passos, A. and
          Cournapeau, D. and Brucher, M. and Perrot, M. and Duchesnay, E.},
  journal={Journal of Machine Learning Research},
  volume={12},
  pages={2825--2830},
  year={2011}
}

@article{Hallin:2021wme,
    author = "Hallin, Anna and Isaacson, Joshua and Kasieczka, Gregor and Krause, Claudius and Nachman, Benjamin and Quadfasel, Tobias and Schlaffer, Matthias and Shih, David and Sommerhalder, Manuel",
    title = "{Classifying anomalies through outer density estimation}",
    eprint = "2109.00546",
    archivePrefix = "arXiv",
    primaryClass = "hep-ph",
    reportNumber = "EFI-20-5, FERMILAB-PUB-21-389-T",
    doi = "10.1103/PhysRevD.106.055006",
    journal = "Phys. Rev. D",
    volume = "106",
    number = "5",
    pages = "055006",
    year = "2022"
}

@article{Raine:2022hht,
    author = "Raine, John Andrew and Klein, Samuel and Sengupta, Debajyoti and Golling, Tobias",
    title = "{CURTAINs for your sliding window: Constructing unobserved regions by transforming adjacent intervals}",
    eprint = "2203.09470",
    archivePrefix = "arXiv",
    primaryClass = "hep-ph",
    doi = "10.3389/fdata.2023.899345",
    journal = "Front. Big Data",
    volume = "6",
    pages = "899345",
    year = "2023"
}

@article{Das:2024eie,
    author = {Das, Ranit and Finke, Thorben and Hein, Marie and Kasieczka, Gregor and Kr{\"a}mer, Michael and M{\"u}ck, Alexander and Shih, David},
    title = "{Accurate and robust methods for direct background estimation in resonant anomaly detection}",
    eprint = "2411.00085",
    archivePrefix = "arXiv",
    primaryClass = "hep-ph",
    reportNumber = "P3H-24-077, TTK-24-45",
    doi = "10.1103/PhysRevD.111.094041",
    journal = "Phys. Rev. D",
    volume = "111",
    number = "9",
    pages = "094041",
    year = "2025"
}

@article{Krause:2023uww,
    author = "Krause, Claudius and Nachman, Benjamin and Pang, Ian and Shih, David and Zhu, Yunhao",
    title = "{Anomaly detection with flow-based fast calorimeter simulators}",
    eprint = "2312.11618",
    archivePrefix = "arXiv",
    primaryClass = "hep-ph",
    reportNumber = "HEPHY-ML-24-03, HEPHY-ML-23-03",
    doi = "10.1103/PhysRevD.110.035036",
    journal = "Phys. Rev. D",
    volume = "110",
    number = "3",
    pages = "035036",
    year = "2024"
}

@article{Matos:2024ggs,
    author = "Matos, Gabriel and Busch, Elena and Park, Ki Ryeong and Gonski, Julia",
    title = "{Semi-supervised permutation invariant particle-level anomaly detection}",
    eprint = "2408.17409",
    archivePrefix = "arXiv",
    primaryClass = "hep-ph",
    doi = "10.1007/JHEP05(2025)116",
    journal = "JHEP",
    volume = "2025",
    number = "05",
    pages = "116",
    year = "2025"
}

@article{Dillon:2022tmm,
    author = "Dillon, Barry M. and Mastandrea, Radha and Nachman, Benjamin",
    title = "{Self-supervised anomaly detection for new physics}",
    eprint = "2205.10380",
    archivePrefix = "arXiv",
    primaryClass = "hep-ph",
    doi = "10.1103/PhysRevD.106.056005",
    journal = "Phys. Rev. D",
    volume = "106",
    number = "5",
    pages = "056005",
    year = "2022"
}

@article{Cowan:2010js,
    author = "Cowan, Glen and Cranmer, Kyle and Gross, Eilam and Vitells, Ofer",
    title = "{Asymptotic formulae for likelihood-based tests of new physics}",
    eprint = "1007.1727",
    archivePrefix = "arXiv",
    primaryClass = "physics.data-an",
    doi = "10.1140/epjc/s10052-011-1554-0",
    journal = "Eur. Phys. J. C",
    volume = "71",
    pages = "1554",
    year = "2011",
    note = "[Erratum: Eur.Phys.J.C 73, 2501 (2013)]"
}

@article{AnomalyDetection,
  author  = {Belis, Vasilis and Odagiu, Patrick and {\AA}rrestad, Thea K.},
  title   = {Machine learning for anomaly detection in particle physics},
  journal = {Reviews in Physics},
  volume  = {12},
  year    = {2024},
  doi     = {10.1016/j.revip.2024.100091},
}

@article{MiningGold,
  author    = {Brehmer, Johann and Louppe, Gilles and Pavez, Juan and Cranmer, Kyle},
  title     = {Mining gold from implicit models to improve likelihood-free inference},
  journal   = {Proceedings of the National Academy of Sciences},
  volume    = {117},
  number    = {10},
  year      = {2020},
  doi       = {10.1073/pnas.1915980117},
}

@article{Huber,
  author  = {Huber, Peter J.},
  title   = {Robust Estimation of a Location Parameter},
  journal = {Annals of Mathematical Statistics},
  year    = {1964},
  volume  = {35},
  number  = {1},
  doi     = {10.1214/aoms/1177703732},
}

@article{Prasanth,
    author = "Matchev, Konstantin T. and Roman, Alexander and Shyamsundar, Prasanth",
    title = "{Uncertainties associated with GAN-generated datasets in high energy physics}",
    eprint = "2002.06307",
    archivePrefix = "arXiv",
    primaryClass = "hep-ph",
    doi = "10.21468/SciPostPhys.12.3.104",
    journal = "SciPost Phys.",
    volume = "12",
    number = "3",
    pages = "104",
    year = "2022"
}

@article{MLreview,
    author = "Butter, Anja and Plehn, Tilman and Schumann, Steffen and 
              Badger, Simon and Caron, Sascha and Cranmer, Kyle and others",
    title = "{Machine learning and LHC event generation}",
    journal = "SciPost Phys.",
    volume = "14",
    number = "4",
    pages = "079",
    eprint = "2203.07460",
    archivePrefix = "arXiv",
    primaryClass = "hep-ph",
    doi = "10.21468/SciPostPhys.14.4.079",
    year = "2023"
}

@article{SBI,
  title   = {The frontier of simulation-based inference},
  author  = {Cranmer, Kyle and Brehmer, Johann and Louppe, Gilles},
  journal = {Proceedings of the National Academy of Sciences},
  volume  = {117},
  number  = {48},
  year    = {2020},
  doi     = {10.1073/pnas.1912789117}
}

@article{Cappelli:2025myc,
    author = "Cappelli, Pietro and Grosso, Gaia and Letizia, Marco and Reyes-Gonz{\'a}lez, Humberto and Zanetti, Marco",
    title = "{Learning to validate generative models: a goodness-of-fit approach}",
    eprint = "2511.09118",
    archivePrefix = "arXiv",
    primaryClass = "stat.ML",
    reportNumber = "TTK-25-36",
    doi = "10.1088/2632-2153/ae73e3",
    journal = "Mach. Learn. Sci. Tech.",
    volume = "7",
    number = "4",
    pages = "045011",
    year = "2026"
}

@article{Grossi:2025pmm,
    author = "Grossi, Samuele and Letizia, Marco and Torre, Riccardo",
    title = "{Comparing generative models with the new physics learning machine}",
    eprint = "2508.02275",
    archivePrefix = "arXiv",
    primaryClass = "stat.ML",
    doi = "10.1016/j.nuclphysb.2026.117349",
    journal = "Nucl. Phys. B",
    volume = "1024",
    pages = "117349",
    year = "2026"
}

@article{Grossi:2024axb,
    author = "Grossi, Samuele and Letizia, Marco and Torre, Riccardo",
    title = "{Refereeing the referees: evaluating two-sample tests for validating generators in precision sciences}",
    eprint = "2409.16336",
    archivePrefix = "arXiv",
    primaryClass = "stat.ML",
    doi = "10.1088/2632-2153/adb3ee",
    journal = "Mach. Learn. Sci. Tech.",
    volume = "6",
    number = "1",
    pages = "015052",
    year = "2025"
}

@article{Bressler:2024wzc,
    author = "Bressler, Shikma and Savoray, Inbar and Zurgil, Yuval",
    title = "{Learning new physics from data: A symmetrized approach}",
    eprint = "2401.09530",
    archivePrefix = "arXiv",
    primaryClass = "hep-ex",
    doi = "10.1103/PhysRevD.110.095004",
    journal = "Phys. Rev. D",
    volume = "110",
    number = "9",
    pages = "095004",
    year = "2024"
}

@article{DAgnolo:2018cun,
    author = "D'Agnolo, Raffaele Tito and Wulzer, Andrea",
    title = "{Learning New Physics from a Machine}",
    eprint = "1806.02350", archivePrefix = "arXiv", primaryClass = "hep-ph",
    doi = "10.1103/PhysRevD.99.015014",
    journal = "Phys. Rev. D", volume = "99", number = "1", pages = "015014",
    year = "2019"
}

@article{dAgnolo:2021aun,
    author = "D'Agnolo, Raffaele Tito and Grosso, Gaia and Pierini, Maurizio and Wulzer, Andrea and Zanetti, Marco",
    title = "{Learning new physics from an imperfect machine}",
    eprint = "2111.13633",
    archivePrefix = "arXiv",
    primaryClass = "hep-ph",
    doi = "10.1140/epjc/s10052-022-10226-y",
    journal = "Eur. Phys. J. C",
    volume = "82",
    number = "3",
    pages = "275",
    year = "2022"
}

@article{LearningMultivariateNP,
  author  = {D'Agnolo, Raffaele Tito and Grosso, Gaia and Pierini, Maurizio and Wulzer, Andrea and Zanetti, Marco},
  title   = {{Learning multivariate new physics}},
  journal = {Eur. Phys. J. C},
  volume  = {81},
  pages   = {89},
  year    = {2021},
  doi     = {10.1140/epjc/s10052-021-08853-y},
  eprint  = {1912.12155},
  archivePrefix = {arXiv},
  primaryClass  = {hep-ph}
}

@article{NonParametricMethods,
  title   = {Learning new physics efficiently with nonparametric methods},
  author  = {Letizia, Marco and Losapio, Gianvito and Rando, Marco and Grosso, Gaia and Wulzer, Andrea and Pierini, Maurizio and Zanetti, Marco and Rosasco, Lorenzo},
  journal = {Eur. Phys. J. C},
  volume  = {82},
  pages   = {879},
  year    = {2022},
  doi     = {10.1140/epjc/s10052-022-10830-y},
  number = {10},
  eprint  = {2204.02317},
  archivePrefix = {arXiv},
  primaryClass  = {hep-ph}
}

@article{NeymanPearsonGoF,
  title   = {{Goodness of fit by Neyman--Pearson testing}},
  author  = {Grosso, Gaia and Letizia, Marco and Pierini, Maurizio and Wulzer, Andrea},
  journal = {SciPost Phys.},
  volume  = {16},
  pages   = {123},
  year    = {2024},
  doi     = {10.21468/SciPostPhys.16.5.123},
  eprint  = {2305.14137},
  archivePrefix = {arXiv},
  primaryClass  = {hep-ph}
}

@article{Grosso:2024wjt,
    author = "Grosso, Gaia and Letizia, Marco",
    title = "{Multiple testing for signal-agnostic searches for new physics with machine learning}",
    eprint = "2408.12296",
    archivePrefix = "arXiv",
    primaryClass = "hep-ph",
    doi = "10.1140/epjc/s10052-024-13722-5",
    journal = "Eur. Phys. J. C",
    volume = "85",
    number = "1",
    pages = "4",
    year = "2025"
}

@article{White1982,
  author  = {White, Halbert},
  title   = {Maximum Likelihood Estimation of Misspecified Models},
  journal = {Econometrica},
  volume  = {50},
  number  = {1},
  pages   = {1--25},
  year    = {1982},
  doi     = {10.2307/1912526},
  url     = {https://www.jstor.org/stable/1912526}
}

@article{BayesianFlows,
  author    = {Bellagente, Marco and Hau{\ss}mann, Manuel and Luchmann, Michel and Plehn, Tilman},
  title     = {Understanding Event-Generation Networks via Uncertainties},
  journal   = {SciPost Physics},
  volume    = {13},
  pages     = {003},
  year      = {2022},
  doi       = {10.21468/SciPostPhys.13.1.003},
  eprint    = {2104.04543},
  archivePrefix = {arXiv},
  primaryClass  = {hep-ph}
}

@inproceedings{Dropout,
  title     = {{Controlled Dropout for Uncertainty Estimation}},
  author    = {Hasan, Mehedi and Hossain, Ibrahim and Rahman, Ashikur and Nahavandi, Saeid},
  booktitle = {2023 IEEE International Conference on Systems, Man, and Cybernetics (SMC)},
  pages     = {973--980},
  year      = {2023},
  doi       = {10.1109/SMC53992.2023.10394101}
}

@article{Sean,
    author = "Benevedes, Sean and Thaler, Jesse",
    title = "{Frequentist uncertainties on neural density ratios with $w_i\,f_i$ ensembles}",
    eprint = "2506.00113",
    archivePrefix = "arXiv",
    primaryClass = "hep-ph",
    reportNumber = "MIT-CTP/5874",
    doi = "10.1103/w28w-x5wh",
    journal = "Phys. Rev. D",
    volume = "112",
    number = "5",
    pages = "056024",
    year = "2025"
}

@article{SPARKER,
    author = "Grosso, Gaia and Hindupur, Sai Sumedh R. and Fel, Thomas and 
              Bright-Thonney, Samuel and Harris, Philip and Ba, Demba",
    title = "{Sparse, self-organizing ensembles of local kernels detect rare statistical anomalies}",
    eprint = "2511.03095",
    journal = "",
    archivePrefix = "arXiv",
    primaryClass = "cs.LG",
    year = "2025"
}

@article{NormalizingFlows,
    author = "Kobyzev, Ivan and Prince, Simon J. D. and Brubaker, Marcus A.",
    title = "{Normalizing flows: an introduction and review of current methods}",
    eprint = "1908.09257",
    archivePrefix = "arXiv",
    primaryClass = "stat.ML",
    doi = "10.1109/TPAMI.2020.2992934",
    journal = "IEEE Trans. Pattern Anal. Mach. Intell.",
    volume = "43",
    number = "11",
    pages = "3964--3979",
    year = "2021"
}

@article{DiSipio:2019imz,
    author = "Di Sipio, Riccardo and Faucci Giannelli, Michele and Ketabchi Haghighat, Sana and Palazzo, Serena",
    title = "{DijetGAN: A Generative-Adversarial Network Approach for the Simulation of QCD Dijet Events at the LHC}",
    eprint = "1903.02433",
    archivePrefix = "arXiv",
    primaryClass = "hep-ex",
    doi = "10.1007/JHEP08(2019)110",
    journal = "JHEP",
    volume = "2019",
    number = "08",
    pages = "110",
    year = "2019"
}

@article{GANplifying,
    author = "Butter, Anja and Diefenbacher, Sascha and Kasieczka, Gregor and Nachman, Benjamin and Plehn, Tilman",
    title = "{GANplifying event samples}",
    eprint = "2008.06545",
    archivePrefix = "arXiv",
    primaryClass = "hep-ph",
    doi = "10.21468/SciPostPhys.10.6.139",
    journal = "SciPost Phys.",
    volume = "10",
    number = "6",
    pages = "139",
    year = "2021"
}

@article{Bayesiamplification,
    author = "Bieringer, Sebastian and Diefenbacher, Sascha and Kasieczka, Gregor and Trabs, Mathias",
    title = "{Calibrating Bayesian generative machine learning for Bayesiamplification}",
    eprint = "2408.00838",
    archivePrefix = "arXiv",
    primaryClass = "cs.LG",
    doi = "10.1088/2632-2153/ad9136",
    journal = "Mach. Learn. Sci. Tech.",
    volume = "5",
    number = "4",
    pages = "045044",
    year = "2024"
}

@article{InfoLimitAmplification,
  title   = {An information theoretic limit to data amplification},
  author  = {Watts, S. J. and Crow, L.},
  journal = {Machine Learning: Science and Technology},
  volume  = {6},
  number  = {2},
  pages   = {025046},
  year    = {2025},
  eprint  = {2412.18041},
  archivePrefix = {arXiv},
  primaryClass  = {stat.ML},
  doi = {10.1088/2632-2153/add78d},
  publisher = {IOP Publishing}
}

@article{ForecastingAmplification,
    author = "Bahl, Henning and Diefenbacher, Sascha and Elmer, Nina and Plehn, Tilman and Spinner, Jonas",
    title = "{Forecasting generative amplification}",
    eprint = "2509.08048",
    archivePrefix = "arXiv",
    primaryClass = "hep-ph",
    doi = "10.21468/SciPostPhys.20.5.150",
    journal = "SciPost Phys.",
    volume = "20",
    number = "5",
    pages = "150",
    year = "2026"
}

@article{Butter:2026dvv,
    author = "Butter, Anja and Diefenbacher, Sascha and Plehn, Tilman and
              Vogel, Lorenz",
    title = "{Know What You Don't Flow}",
    eprint = "2608.21509",
    journal = "",
    archivePrefix = "arXiv",
    primaryClass = "hep-ph",
    year = "2026"
}

@article{Calomplification,
    author = "Bieringer, Sebastian and Butter, Anja and Diefenbacher, Sascha and
              Eren, Engin and Gaede, Frank and Hundhausen, Daniel and
              Kasieczka, Gregor and Nachman, Benjamin and Plehn, Tilman and
              Trabs, Mathias",
    title = "{Calomplification --- the power of generative calorimeter models}",
    eprint = "2202.07352", archivePrefix = "arXiv", primaryClass = "hep-ph",
    doi = "10.1088/1748-0221/17/09/P09028",
    journal = "JINST", volume = "17", number = "09", pages = "P09028", year = "2022"
}

@article{Das:2023ktd,
    author = "Das, Ranit and Favaro, Luigi and Heimel, Theo and Krause, Claudius and Plehn, Tilman and Shih, David",
    title = "{How to understand limitations of generative networks}",
    eprint = "2305.16774",
    archivePrefix = "arXiv",
    primaryClass = "hep-ph",
    doi = "10.21468/SciPostPhys.16.1.031",
    journal = "SciPost Phys.",
    volume = "16",
    number = "1",
    pages = "031",
    year = "2024"
}

@article{Kansal:2022spb,
    author = "Kansal, Raghav and Li, Anni and Duarte, Javier and Chernyavskaya, Nadezda and Pierini, Maurizio and Orzari, Breno and Tomei, Thiago",
    title = "{Evaluating generative models in high energy physics}",
    eprint = "2211.10295",
    archivePrefix = "arXiv",
    primaryClass = "hep-ex",
    reportNumber = "FERMILAB-PUB-22-872-CMS-PPD",
    doi = "10.1103/PhysRevD.107.076017",
    journal = "Phys. Rev. D",
    volume = "107",
    number = "7",
    pages = "076017",
    year = "2023"
}

@article{Krause:classifier-metric,
    author = "Krause, Claudius and Shih, David",
    title = "{Fast and accurate simulations of calorimeter showers with normalizing flows}",
    eprint = "2106.05285",
    archivePrefix = "arXiv",
    primaryClass = "physics.ins-det",
    doi = "10.1103/PhysRevD.107.113003",
    journal = "Phys. Rev. D",
    volume = "107",
    number = "11",
    pages = "113003",
    year = "2023"
}

@article{massey1951kolmogorov,
    author = "Massey, Jr., Frank J.",
    title = "{The Kolmogorov--Smirnov test for goodness of fit}",
    doi = "10.1080/01621459.1951.10500769",
    journal = "J. Am. Stat. Assoc.",
    volume = "46", number = "253", pages = "68--78", year = "1951"
}

@inproceedings{Qu:2022mxj,
    author = "Qu, Huilin and Li, Congqiao and Qian, Sitian",
    title = "{Particle Transformer for Jet Tagging}",
        booktitle = "Proceedings of the 39th International Conference on Machine Learning",
    series = "Proceedings of Machine Learning Research",
    volume = "162",
    publisher = "PMLR",
    pages = "18281--18292",
    year = "2022",
    eprint = "2202.03772",
    archivePrefix = "arXiv",
    primaryClass = "hep-ph"
}

@inproceedings{Chen:2020simclr,
    author = "Chen, Ting and Kornblith, Simon and Norouzi, Mohammad and Hinton, Geoffrey",
    title = "{A simple framework for contrastive learning of visual representations}",
    booktitle = "Proceedings of the 37th International Conference on Machine Learning",
    series = "Proceedings of Machine Learning Research",
    volume = "119",
    publisher = "PMLR",
    pages = "1597--1607",
    year = "2020",
    eprint = "2002.05709",
    archivePrefix = "arXiv",
    primaryClass = "cs.LG"
}

@misc{nflows,
    author       = "Durkan, Conor and Bekasov, Artur and Murray, Iain and Papamakarios, George",
    title        = "{nflows: normalizing flows in PyTorch}",
    howpublished = "Zenodo, version v0.14",
    doi          = "10.5281/zenodo.4296287",
    month        = nov,
    year         = "2020"
}

@inproceedings{KingmaBa2015Adam,
    author        = "Kingma, Diederik P. and Ba, Jimmy",
    title         = "{Adam: A method for stochastic optimization}",
    booktitle     = "Proceedings of the 3rd International Conference on Learning Representations (ICLR)",
    year          = "2015",
    eprint        = "1412.6980",
    archivePrefix = "arXiv",
    primaryClass  = "cs.LG"
}

@book{hall1992bootstrap,
    author = "Hall, Peter",
    title = "{The bootstrap and Edgeworth expansion}",
    series = "Springer Series in Statistics",
    publisher = "Springer",
    address = "New York",
    doi = "10.1007/978-1-4612-4384-7",
    isbn = "978-0-387-97720-1",
    year = "1992"
}

@article{Tong:2026,
    author        = "Tong, Shelley and Harris, Philip and Grosso, Gaia",
    title         = "{Covariant Contrastive Learning for Uncertainty-Aware Anomaly Detection}",
    journal = "",
    eprint        = "2609.35755",
    archivePrefix = "arXiv",
    primaryClass  = "hep-ph",
    month         = "9",
    year          = "2026"
}

@article{Haussmann:2026,
    author        = "Hau{\ss}mann, Manuel and Winterhalder, Ramon and Ubiali, Maria",
    title         = "{Uncertainty in Physics and AI: Taxonomy, Quantification, and Validation}",
    journal = "",
    eprint        = "2605.10378",
    archivePrefix = "arXiv",
    primaryClass = "stat.ML",
    year          = "2026"
}

@inproceedings{Jimenez:2026,
    author        = "Jim{\'e}nez, Sebasti{\'a}n and J{\"u}rgens, Mira and Waegeman, Willem",
    title         = "{Position: Epistemic uncertainty estimation methods are fundamentally incomplete}",
    booktitle     = "Proceedings of the 43rd International Conference on Machine Learning",
    series        = "Proceedings of Machine Learning Research",
    volume        = "306",
    publisher     = "PMLR",
    pages         = "170691--170717",
    year          = "2026",
    eprint        = "2505.23506",
    archivePrefix = "arXiv",
    primaryClass  = "cs.LG"
}

@article{Harris:2024sra,
    author = "Harris, Philip and Krupa, Jeffrey and Kagan, Michael and Maier, Benedikt and Woodward, Nathaniel",
    title = "{Resimulation-based self-supervised learning for pretraining physics foundation models}",
    eprint = "2403.07066",
    archivePrefix = "arXiv",
    primaryClass = "hep-ph",
    doi = "10.1103/PhysRevD.111.032010",
    journal = "Phys. Rev. D",
    volume = "111",
    number = "3",
    pages = "032010",
    year = "2025"
}

@article{Frate:2017mai,
    author = "Frate, Meghan and Cranmer, Kyle and Kalia, Saarik and Vandenberg-Rodes, Alexander and Whiteson, Daniel",
    title = "{Modeling Smooth Backgrounds and Generic Localized Signals with Gaussian Processes}",
    eprint = "1709.05681",
    journal = "",
    archivePrefix = "arXiv",
    primaryClass = "physics.data-an",
    month = "9",
    year = "2017"
}

@article{ATLAS:2026ugi,
    author = "Aad, Georges and others",
    collaboration = "ATLAS",
    title = "{Search for new scalars via $X \rightarrow SH \rightarrow b\bar{b}b\bar{b}$ in proton-proton collisions at $\sqrt{s} = 13$ TeV with the ATLAS detector}",
    eprint = "2607.18484",
    journal = "",
    archivePrefix = "arXiv",
    primaryClass = "hep-ex",
    reportNumber = "CERN-EP-2026-202",
    month = "7",
    year = "2026"
}

@book{McLachlanPeel,
    author    = "McLachlan, Geoffrey and Peel, David",
    title     = "{Finite Mixture Models}",
    publisher = "John Wiley \& Sons",
    address   = "New York",
    series    = "Wiley Series in Probability and Statistics",
    year      = "2000",
    doi       = "10.1002/0471721182"
}

@article{Menary:2021,
    author        = "Menary, Stephen B. and Price, Darren D.",
    title         = "{Learning to discover: expressive Gaussian mixture models for multi-dimensional simulation and parameter inference in the physical sciences}",
    eprint        = "2108.11481",
    archivePrefix = "arXiv",
    primaryClass  = "physics.data-an",
    doi           = "10.1088/2632-2153/ac4a3b",
    journal       = "Mach. Learn. Sci. Tech.",
    volume        = "3",
    number        = "1",
    pages         = "015021",
    year          = "2022"
}

@article{Bouveyron:2014,
    author  = "Bouveyron, Charles and Brunet-Saumard, Camille",
    title   = "{Model-based clustering of high-dimensional data: A review}",
    doi     = "10.1016/j.csda.2012.12.008",
    journal = "Comput. Stat. Data Anal.",
    volume  = "71",
    pages   = "52--78",
    year    = "2014"
}

@article{dempster1977maximum,
    author = "Dempster, Arthur P. and Laird, Nan M. and Rubin, Donald B.",
    title = "{Maximum likelihood from incomplete data via the EM algorithm}",
    doi = "10.1111/j.2517-6161.1977.tb01600.x",
    journal = "J. Roy. Statist. Soc. Ser. B",
    volume = "39",
    number = "1",
    pages = "1--22",
    year = "1977"
}

@article{redner1984mixture,
    author = "Redner, Richard A. and Walker, Homer F.",
    title = "{Mixture densities, maximum likelihood and the EM algorithm}",
    doi = "10.1137/1026034",
    journal = "SIAM Rev.",
    volume = "26",
    number = "2",
    pages = "195--239",
    year = "1984"
}

@article{louis1982finding,
    author = "Louis, Thomas A.",
    title = "{Finding the observed information matrix when using the EM algorithm}",
    doi = "10.1111/j.2517-6161.1982.tb01203.x",
    journal = "J. Roy. Statist. Soc. Ser. B",
    volume = "44", number = "2", pages = "226--233", year = "1982"
}

@article{mclachlan2019finite,
    author = "McLachlan, Geoffrey J. and Lee, Sharon X. and Rathnayake, Suren I.",
    title = "{Finite mixture models}",
    doi = "10.1146/annurev-statistics-031017-100325",
    journal = "Ann. Rev. Stat. Appl.",
    volume = "6",
    number = "1",
    pages = "355--378",
    year = "2019"
}

@book{mclachlan2008algorithm,
    author = "McLachlan, Geoffrey J. and Krishnan, Thriyambakam",
    title = "{The EM algorithm and extensions}",
    edition = "2nd",
    series = "Wiley Series in Probability and Statistics",
    publisher = "Wiley", address = "Hoboken, NJ",
    doi = "10.1002/9780470191613",
    isbn = "978-0-471-20170-0",
    year = "2008"
}

@article{yakowitz1968identifiability,
    author = "Yakowitz, Sidney J. and Spragins, John D.",
    title = "{On the identifiability of finite mixtures}",
    doi = "10.1214/aoms/1177698520",
    journal = "Ann. Math. Statist.",
    volume = "39",
    number = "1",
    pages = "209--214",
    year = "1968"
}

@article{lilliefors1967kolmogorov,
    author = "Lilliefors, Hubert W.",
    title = "{On the Kolmogorov--Smirnov test for normality with mean and variance unknown}",
    doi = "10.1080/01621459.1967.10482916",
    journal = "J. Am. Stat. Assoc.",
    volume = "62",
    number = "318",
    pages = "399--402",
    year = "1967"
}

@article{durbin1973weak,
    author = "Durbin, James",
    title = "{Weak convergence of the sample distribution function when parameters are estimated}",
    doi = "10.1214/aos/1176342365",
    journal = "Ann. Statist.",
    volume = "1",
    number = "2",
    pages = "279--290",
    year = "1973"
}

@article{stute1993bootstrap,
    author = "Stute, Winfried and Gonz{\'a}lez Manteiga, Wenceslao and Presedo Quindimil, Manuel",
    title = "{Bootstrap based goodness-of-fit-tests}",
    doi = "10.1007/BF02613687",
    journal = "Metrika",
    volume = "40",
    number = "1",
    pages = "243--256",
    year = "1993"
}

@article{babu2004goodness,
    author = "Babu, G. Jogesh and Rao, Calyampudi R.",
    title = "{Goodness-of-fit tests when parameters are estimated}",
    journal = "Sankhy{\=a}",
    volume = "66",
    number = "1",
    pages = "63--74",
    year = "2004"
}

@article{genest2008validity,
    author = "Genest, Christian and R{\'e}millard, Bruno",
    title = "{Validity of the parametric bootstrap for goodness-of-fit testing in semiparametric models}",
    doi = "10.1214/07-AIHP148",
    journal = "Ann. Inst. H. Poincar{\'e} Probab. Statist.",
    volume = "44",
    number = "6",
    pages = "1096--1127",
    year = "2008"
}

@article{hayes1989application,
    author = "Hayes, Kenneth G. and Perl, Martin L. and Efron, Bradley",
    title = "{Application of the bootstrap statistical method to the tau-decay-mode problem}",
    doi = "10.1103/PhysRevD.39.274",
    journal = "Phys. Rev. D",
    volume = "39",
    number = "1",
    pages = "274",
    year = "1989"
}

@book{cowan1998statistical,
    author = "Cowan, Glen",
    title = "{Statistical data analysis}",
    publisher = "Clarendon Press",
    address = "Oxford",
    isbn = "978-0-19-850156-5",
    year = "1998"
}

@article{Grosso:2026fgo,
    author = "Grosso, Gaia and Mikuni, Vinicius and Heinrich, Lukas",
    title = "{Are We Ready for AI-Driven Discovery? AI Verification Before the Next Fundamental Physics Breakthrough}",
    eprint = "2607.10039",
    journal = "",
    archivePrefix = "arXiv",
    primaryClass = "physics.data-an",
    month = {7},
    year = {2026}
}

@inproceedings{gal2016dropout,
  title={Dropout as a {B}ayesian approximation: Representing model uncertainty in deep learning},
  author={Gal, Yarin and Ghahramani, Zoubin},
  booktitle={International Conference on Machine Learning},
  pages={1050--1059},
  year={2016},
  organization={PMLR}
}

@inproceedings{immer2021improving,
    author = "Immer, Alexander and Korzepa, Maciej and Bauer, Matthias",
    title = "{Improving predictions of Bayesian neural nets via local linearization}",
    booktitle = "Proceedings of the 24th International Conference on Artificial Intelligence and Statistics",
    series = "Proceedings of Machine Learning Research",
    volume = "130", publisher = "PMLR",
    pages = "703--711", year = "2021",
    eprint = "2008.08400", archivePrefix = "arXiv", primaryClass = "stat.ML"
}

@inproceedings{daxberger2021laplace,
    author = "Daxberger, Erik and Kristiadi, Agustinus and Immer, Alexander and Eschenhagen, Runa and Bauer, Matthias and Hennig, Philipp",
    title = "{Laplace Redux -- Effortless Bayesian Deep Learning}",
    booktitle = "Advances in Neural Information Processing Systems",
    volume = "34", pages = "20089--20103", year = "2021",
    eprint = "2106.14806", archivePrefix = "arXiv", primaryClass = "cs.LG"
}

@inproceedings{Khosla:2020supcon,
    author = "Khosla, Prannay and Teterwak, Piotr and Wang, Chen and
              Sarna, Aaron and Tian, Yonglong and Isola, Phillip and
              Maschinot, Aaron and Liu, Ce and Krishnan, Dilip",
    title = "{Supervised contrastive learning}",
    booktitle = "Advances in Neural Information Processing Systems",
    volume = "33", pages = "18661--18673", year = "2020",
    eprint = "2004.11362", archivePrefix = "arXiv", primaryClass = "cs.LG"
}

@dataset{reissel_2026_23693193,
  author       = {Reissel, Christina and
                  Badaracco, Giada and
                  Benevedes, Sean and
                  Grosso, Gaia and
                  Harris, Philip},
  title        = {4D SupCon JetClass Embedding (QCD class)},
  month        = oct,
  year         = 2026,
  publisher    = {Zenodo},
  doi          = {10.5281/zenodo.23693193},
  url          = {https://doi.org/10.5281/zenodo.23693193},
}

\end{document}